\pdfoutput=1

\documentclass[11pt]{article}

\usepackage[]{ACL2023}

\usepackage{times}
\usepackage{latexsym}

\usepackage[T1]{fontenc}

\usepackage[utf8]{inputenc}

\usepackage{microtype}

\usepackage{inconsolata}

\usepackage[breaklinks]{hyperref}
\usepackage{xcolor}
\hypersetup{
    colorlinks,
    linkcolor={blue!80!black},
    citecolor={blue!50!black},
    urlcolor={blue!80!black}
}
\usepackage{url}

\usepackage{array}
\newcolumntype{P}[1]{>{\centering\arraybackslash}p{#1}}
\usepackage{algorithm,algorithmic}
\usepackage{wrapfig}
\usepackage{bbm}
\usepackage{caption}
\usepackage{paralist}
\usepackage{xspace}
\newcommand{\ourmetric}{\textsc{Rev}\xspace}

\usepackage{trimclip}

\newcommand{\gold}{\textsc{Y$^*$;R$^*$}\xspace}
\newcommand{\base}{\textsc{Y$^*$;B}\xspace}
\newcommand{\outputr}{\textsc{XY$^*$\textrightarrow R}\xspace}
\newcommand{\outputyr}{\textsc{X\textrightarrow YR}\xspace}
\newcommand{\outputry}{\textsc{X\textrightarrow RY}\xspace}

\usepackage{comment}

\usepackage{fdsymbol}

\newcommand*{\affaddr}[1]{#1} 
\newcommand*{\affmark}[1][*]{\textsuperscript{#1}}
\newcommand*{\email}[1]{\small\texttt{#1}}

\usepackage[]{trackchanges}
\addeditor{SS}
\addeditor{fb}
\addeditor{XR}
\addeditor{YJ}
\addeditor{HC}
\addeditor{YC}

\usepackage[linecolor=orange,size=tiny]{todonotes}

\usepackage{makecell}

\usepackage{amsmath,amsfonts,bm} 
\usepackage{pifont} 
\usepackage{graphicx,subfigure,epsfig,fancybox} 
\usepackage{float}
\usepackage{color} 
\usepackage{multirow}
\usepackage{natbib}

\usepackage{tikz}
\usetikzlibrary{shapes.geometric, positioning, calc}
\usetikzlibrary{arrows,shapes,calc}
\usetikzlibrary{trees,positioning,mindmap,shadows,fit}
\usetikzlibrary{decorations.pathreplacing}
\usetikzlibrary{tikzmark} 
\usetikzlibrary{intersections} 

\newcommand{\revised}[1]{\textcolor{blue}{}}

\usepackage{adjustbox}
\usepackage{array}
\usepackage{booktabs}

\newcolumntype{R}[2]{%
    >{\adjustbox{angle=#1,lap=\width-(#2)}\bgroup}%
    l%
    <{\egroup}%
}

\usepackage{setspace}

\title{\ourmetric: Information-Theoretic Evaluation of Free-Text Rationales}
\author{%
 Hanjie Chen\affmark[$\heartsuit$]\thanks{~~Work done during an internship at AI2.}\quad 
 Faeze Brahman\affmark[$\spadesuit$$\diamondsuit$]\quad 
 Xiang Ren\affmark[$\spadesuit$$\clubsuit$]\quad
 Yangfeng Ji\affmark[$\heartsuit$]\\
 \textbf{Yejin Choi\affmark[$\spadesuit$$\diamondsuit$]\quad 
 Swabha Swayamdipta\affmark[$\clubsuit$]} \\
\affaddr{\affmark[$\heartsuit$]Department of Computer Science, University of Virginia} \\
\affaddr{\affmark[$\spadesuit$]Allen Institute for AI}\quad
\affaddr{\affmark[$\clubsuit$]University of Southern California}\\
\affaddr{\affmark[$\diamondsuit$]
Paul G. Allen School of Computer Science \& Engineering, University of Washington}\\
	\email{\{hc9mx,yangfeng\}@virginia.edu}\qquad\email{\{faezeb,xiangr,yejinc\}@allenai.org}\qquad\email{swabhas@usc.edu}
}

\begin{document}
\maketitle

\begin{abstract}
Generating free-text rationales is a promising step towards explainable NLP, yet evaluating such rationales remains a challenge. 
Existing metrics have mostly focused on measuring the association between the rationale and a given label. 
We argue that an ideal metric should focus on the new information uniquely provided in the rationale that is otherwise not provided in the input or the label. 
We investigate this research problem from an information-theoretic perspective using conditional $\mathcal{V}$-information \citep{hewitt-etal-2021-conditional}.  
More concretely, we propose a metric called \ourmetric (\underline{R}ationale \underline{E}valuation with conditional \underline{$\mathcal{V}$}-information), to quantify the amount of new, label-relevant information in a rationale \textit{beyond} the information already available in the input or the label. 
Experiments across four benchmarks with reasoning tasks, including chain-of-thought, demonstrate the effectiveness of \ourmetric in evaluating rationale-label pairs, compared to existing metrics. 
We further demonstrate \ourmetric is consistent with human judgments on rationale evaluations and provides more sensitive measurements of new information in free-text rationales.
When used alongside traditional performance metrics, \ourmetric provides deeper insights into models' reasoning and prediction processes.\footnote{Our code is publicly available at \url{https://github.com/HanjieChen/REV}}
\end{abstract}

\section{Introduction}
\label{sec:intro} 

\begin{figure*}[ht]
  \centering
  \includegraphics[width=0.95\textwidth]{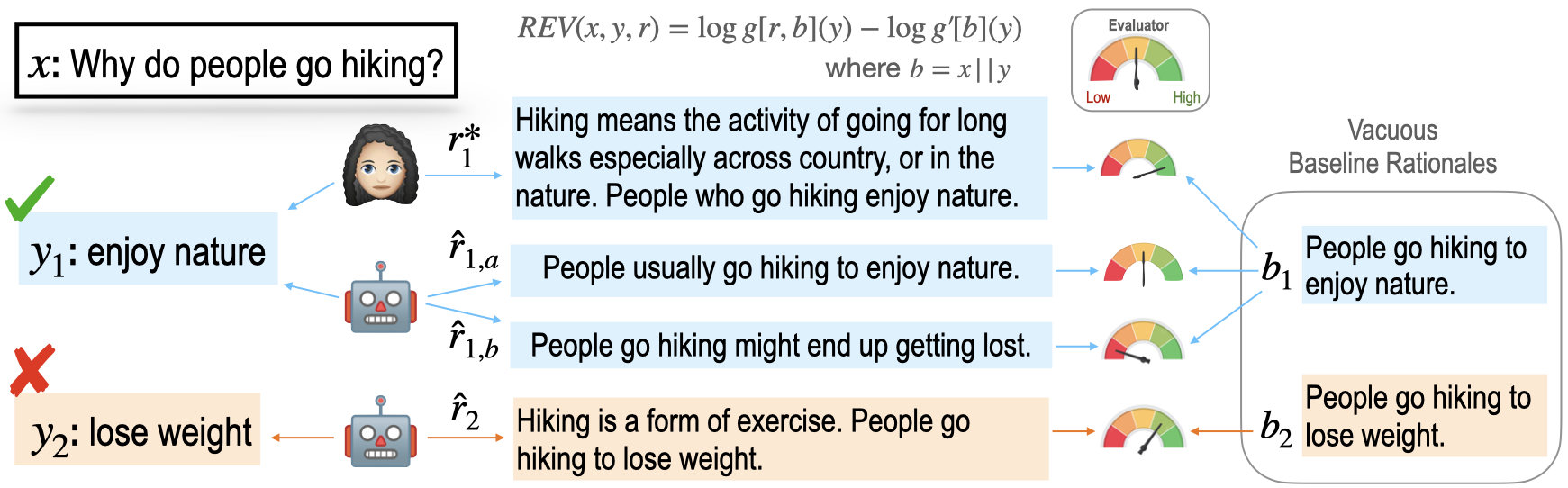}
  \caption{\label{fig:illustrations} Our evaluation framework for different free-text rationales ($r$).
  ${r}_1^*$ is a human-written rationale, ${\hat{r}}_{1,a}$ and ${\hat{r}}_{1,b}$ are two generated rationales for the true label $y_1$.
  Our metric, \ourmetric, based on CVI \citep{hewitt-etal-2021-conditional} is able to distinguish all three rationales by measuring how much new and label-relevant information each adds over a vacuous rationale, ${b}$; performance-based evaluations can only distinguish between ${\hat{r}}_{1,a}$ and ${\hat{r}}_{1,b}$.
  For an (arguably) incorrect label, $y_2$, \ourmetric still gives a positive score highlighting that ${\hat{r}}_2$ is able to provide new information for why it supports $y_2$.
  Prediction accuracy can be augmented with \ourmetric to provide a fuller interpretability of model decisions.
  }
\end{figure*}

Model explanations have been indispensable for trust and interpretability in natural language processing (NLP) \citep{ribeiro2016should, ribeiro-etal-2020-beyond, lipton2018mythos, chen-etal-2020-generating, chen-etal-2021-explaining}. 
Free-text rationales, which explain a model prediction in natural language, have been especially appealing due to their flexibility in eliciting the reasoning process behind the model's decision making \citep{camburu2018snli, narang2020wt5, rajani2019explain, kumar2020nile, brahman2021learning}, making them closer to human explanations.
However, existing metrics for free-text rationale evaluation remain narrowly focused on the extent to which a rationale can help a (proxy) model predict the label it explains (i.e., accuracy based) \citep{hase-etal-2020-leakage, wiegreffe-etal-2021-measuring}. 
These metrics offer little understanding of the \textit{new information} contained in the rationale, as added to the original input, that could \textit{explain why the label is selected}---the very purpose a rationale is designed to serve.
For instance, the two rationales $r_1^*$ and $\hat r_{1,a}$ in Fig. \ref{fig:illustrations} would be considered equally valuable under existing metrics, even though they supply different amount of novel and relevant information.

In this paper, we overcome this shortcoming by introducing an automatic evaluation for free-text rationales along two dimensions: (1) whether the rationale supports (i.e., is predictive of) the intended label, and (2) how much \textit{new information} does it provide to justify the label, \textbf{beyond} what is contained in the input.
For example, rationale $\hat r_{1,b}$ in Fig. \ref{fig:illustrations} violates (1) because it is not predictive of the label, ``\texttt{enjoy nature}''. 
Rationale $\hat r_{1,a}$ does support the label but contains no new information that justifies it, \textit{beyond} what is stated in the input $x$; thus, it violates (2). 
Rationale $r_{1}^*$ is satisfied along both dimensions: it supports the label and does so by providing new and relevant information, beyond what is in the input. 
Our proposed evaluation is designed to penalize both $\hat r_{1,a}$ and $\hat r_{1,b}$, while rewarding rationales like $r_{1}^*$.

We introduce \ourmetric\footnote{For \underline{R}ationale \underline{E}valuation with conditional \underline{$\mathcal{V}$}-information.}, which adapts an information-theoretic framework from \citet{xu2019theory} for evaluating free-text rationales along the two dimensions mentioned above.
Specifically, \ourmetric is based on conditional $\mathcal{V}$-information \citep{hewitt-etal-2021-conditional}, which quantifies the degree of information contained in a representation \textit{beyond} another (baseline) representation, accessible to a model family $\mathcal{V}$.
As our baseline representation, we consider any vacuous rationale which simply (and declaratively) combines an input with a given label, without providing any new information relevant to answering why the label was chosen.
\ourmetric adapts conditional $\mathcal{V}$-information to evaluate rationales, where we compare two representations---one from an evaluation model trained to produce the label given the input and the rationale, and the other from another evaluation model for the same task but considering only the input (disguised as a vacuous rationale). 
Other metrics do not take into consideration vacuous rationales, and are hence unable to measure new and label-relevant information in rationales.

In our experiments, we present evaluations with \ourmetric for rationales under two reasoning tasks, commonsense question-answering (CQA; \citealp{talmor2018commonsenseqa}) and natural language inference (NLI; \citealp{bowman-etal-2015-large}), across four benchmarks. 
Several quantitative evaluations demonstrate the capabilities of \ourmetric in providing evaluations along new dimensions for free-text rationales, while also being more consistent with human judgements compared to existing metrics.
We also provide comparisons to demonstrate the sensitivity of \ourmetric to various degrees of input perturbations.
Additionally, evaluation with \ourmetric offers insights into why rationales obtained through chain-of-thought prompting \citep{wei2022chain} do not necessarily improve prediction performance. 

\section{\ourmetric: Information-Theoretic Evaluation of Rationales}
\label{sec:method}

We introduce a new metric, \ourmetric, \underline{R}ationale \underline{E}valuation with conditional \underline{$\mathcal{V}$}-information, for evaluation of free-text rationales on the proposed dimensions (\S\ref{sec:compute_rev}), based on the framework of conditional $\mathcal{V}$-information (\S\ref{sec:rev}).

We consider the setting where we have input $X \in \mathcal{X}$, label $Y\in \mathcal{Y}$, and free-text rationale $R\in \mathcal{R}$ generated for label $Y$. 
A common strategy to evaluate rationale $R$ is through an evaluator function $f: Z \rightarrow Y$, which maps a variable $Z$ to a label distribution.
Here, $Z$ can be defined based on the evaluation framework;  e.g., $Z$ can be a concatenation of $X$ and $R$, or contains only $X$.
These metrics evaluate the utility of $R$ based on how much $R$ helps $f$ predict $Y$. 
The evaluator $f$ is typically trained on a set of input, label and rationale triples $\mathcal{D}_{\text{train}}=\{(x_j, y_j, r_j)\}$, and applied to $\mathcal{D}_{\text{test}}=\{(x_i,y_i,r_i)\}$ for evaluation.
The utility of $R$ is formulated as the difference between the performance of the evaluator on predicting $Y$ with $R$, and without it, i.e. 
\begin{equation}
	\label{eq:diff_perf}
	\text{Perf}[f(Y|X,R)] - \text{Perf}[f(Y|X)],
\end{equation}
where a larger performance gap indicates a better rationale. 
Existing metrics \citep{hase-etal-2020-leakage, wiegreffe-etal-2021-measuring} compute the performance gap based on prediction accuracies. 

However, accuracy-based evaluation can only indicate whether or not a rationale is predictive of a label, but cannot quantify how much \textit{new information the rationale provides to justify the label}. 
Figure \ref{fig:illustrations} illustrates this issue via an example. 
Here, accuracy-based evaluation can distinguish between ${\hat{r}}_{1,a}$ and ${\hat{r}}_{1,b}$ since ${\hat{r}}_{1,a}$ supports $y_1$ and ${\hat{r}}_{1,b}$ does not. 
However, it is unable to distinguish between ${r}_1^*$ and ${\hat{r}}_{1,a}$ (since both are predictive of $y_1$), despite the fact that ${\hat{r}}_{1,a}$ does not provide any unique and relevant information to answer why the label should be $y_1$. 
In practice, vacuous rationales such as ${\hat{r}}_{1,a}$ are commonly seen in model generations \citep{sun2022investigating,wiegreffe2021teach}. 
This calls for an evaluation metric which is able to identify and penalize such vacuous rationales.

\subsection{An Information-Theoretic Perspective on Rationale Evaluation}
\label{sec:rev}

The key quantity of interest for our evaluation of rationale $R$ is the amount of new information expressed in $R$ (e.g., background knowledge, reasoning process) that can justify a label $Y$.
The mutual information between $R$ and $Y$, $I(Y;R)$, can be helpful for evaluating this quantity.
However, we are not interested in the information that is already captured in the input $X$.
A \textbf{vacuous} rationale, such as ${\hat{r}}_{1,a}$ in Fig. \ref{fig:illustrations}---which simply combines the input $X$ and the label, $Y$ declaratively---captures all the information in $X$ and $Y$ without specifying any new information to help understand why $Y$ has been chosen for $X$. 
We denote such rationales as $B$. 
Thus, we argue that a good evaluation metric must be able to measure the amount of new and label-relevant information contained in a rationale \textit{beyond} what is contained in any vacuous rationale, $B$, that leads to the prediction of $Y$. 
Then the new information in $R$ beyond what is available in $B$ can be grounded with conditional mutual information \citep{shannon1948mathematical} as follows,
\begin{equation}
	\label{eq:cond_info}
	I(Y;R\mid B)=I(Y;R, B)-I(Y;B),
\end{equation}
where the difference of two information quantities demonstrates the performance gap in \autoref{eq:diff_perf}.

Directly computing mutual information, however, is challenging because true distributions of random variables are usually unknown, and we do not have unbounded computation.
A recently introduced information-theoretic framework called $\mathcal{V}$-information circumvents this by restricting the computation to certain predictive model families, $\mathcal{V}$ \citep{xu2019theory}.
Given a model family $\mathcal{V}$ that maps two random variables $R$ and $Y$, $\mathcal{V}$-information defines the usable information that can be extracted from $R$ by models in $\mathcal{V}$ to predict $Y$, i.e. $I_{\mathcal{V}}(R \rightarrow Y)$. 
If $\mathcal{V}$ generalizes to the set of all possible functions, then $\mathcal{V}$-information is mutual information \citep{shannon1948mathematical}. 
In practice, it is feasible to estimate the usable information from $R$ about $Y$ by selecting any neural model without frozen parameters as $\mathcal{V}$.\footnote{
Please see \citet{xu2019theory} for a detailed discussion of properties such as optional ignorance that a predictive family $\mathcal{V}$ must follow.
}
Our approach to evaluate rationales builds on a modification of this framework for conditional information by  \citet{hewitt-etal-2021-conditional}, as described below.

\paragraph{Conditional $\mathcal{V}$-information}

Following conditional mutual information in information theory \citep{cover2006elements}, $\mathcal{V}$-information has been extended to conditional $\mathcal{V}$-information (CVI; \citealp{hewitt-etal-2021-conditional}). 
CVI quantifies the $\mathcal{V}$-usable information in $R$ about $Y$ conditioned on a variable $B$, i.e.
\begin{equation*}
	\label{eq:cvi}
	I_{\mathcal{V}}(R \rightarrow Y \mid B) = H_{\mathcal{V}}(Y\mid B) - H_{\mathcal{V}}(Y \mid R, B).
\end{equation*}
Here $B$ is any vacuous rationale that leads to the prediction of $Y$. 
In this work, we consider $B$ simply as the declarative combination of $X$ and $Y$. 
$H_{\mathcal{V}}(\cdot \mid \cdot)$ is the conditional $\mathcal{V}$-entropy \citep{xu2019theory, hewitt-etal-2021-conditional, ethayarajh2022understanding}, defined as 
\begin{align}
    H_{\mathcal{V}}(Y\mid B)&=\inf_{f \in \mathcal{V}} \mathbb{E}[-\log f[b](y)] \label{eq:entropy_1} \\
    H_{\mathcal{V}}(Y\mid R, B)&=\inf_{f \in \mathcal{V}} \mathbb{E}[-\log f[r, b](y)], \label{eq:entropy_2}
\end{align}
where $f[b]$ and $f[r,b]$ produce a probability distribution over the labels given $b$ and $[r,b]$ as inputs respectively.\footnote{$[r, b]$ is the concatenation of $r$ and $b$. Please see Appendix \ref{sec:property_cvi} for further details on CVI.} 
Further, given $g', g \in \mathcal{V}$ which optimize Equations \ref{eq:entropy_1} and \ref{eq:entropy_2} respectively, 
we consider pointwise CVI for individual triples $(r,y,b)$: 
\begin{equation}
\label{eq:pcvi}
- \log g'[b](y) + \log g[r, b](y).
\end{equation}

\subsection{Computing \ourmetric for Rationale Evaluation}
\label{sec:compute_rev}

Building on the framework of CVI, we propose a new metric \ourmetric, for \underline{R}ationale \underline{E}valuation with conditional \underline{$\mathcal{V}$}-information.
We compute \ourmetric over a given test set, $\mathcal{D}_\text{test}=\{(x_i, y_i, r_i)\}$, by estimating CVI over the set with evaluation models, $g, g' \in \mathcal{V}$. 
For a test example $(x, y, r)$, the \ourmetric score denoted as $\text{\ourmetric}(x, y, r)$ is computed based on \autoref{eq:pcvi}, where $b$ is constructed by combining $x$ and $y$. 
, 
\begin{equation}
    \nonumber
	\ourmetric(x, y, r) = - \log g'[b](y) + \log g[r, b](y).
\end{equation}
The \ourmetric score for the entire test corpus $\mathcal{D}_{\text{test}}$, is given by the average pointwise \ourmetric score: 
\begin{equation}
	\label{eq:compute_rev}
	\ourmetric_\mathcal{D}=\frac{1}{|\mathcal{D}_{\text{test}}|}\sum_{i=1}^{|\mathcal{D}_{\text{test}}|} \ourmetric(x_i, y_i, r_i).
\end{equation}

\begin{table*}[t]
\small
	\centering
	\begin{tabular}{P{1cm}P{5.5cm}P{2cm}P{5.5cm}}
		\toprule
		Task & Input & Label & Vacuous Baseline Rationale \\
		\midrule
		CQA  & Where can personal mushrooms be kept fresh? & refrigerator & Personal mushrooms can be kept fresh in the refrigerator. \\
		\midrule
		NLI & Premise: A dog running in the surf. Hypothesise: A dog is at the beach. & entailment & A dog running in the surf indicates a dog is at the beach. \\
		\bottomrule
	\end{tabular}
	\caption{Examples of constructed vacuous baseline rationales for CQA and NLI tasks. For NLI, the vacuous baseline rationale was obtained after paraphrasing.}
	\label{tab:construct_rat}
\end{table*}
\begin{algorithm}[H]
  \caption{Computing \ourmetric Scores}
  \label{alg:compute_rev}
  \begin{algorithmic}[1]
                \STATE {\bf Input}: evaluation models $g$ and $g'$, test set $\mathcal{D}_{\text{test}}=\{(x_i, y_i, r_i)\}$
    		\STATE Initialize an empty list $\mathcal{S}$ 
    		\FOR {$(x_i, y_i, r_i) \in \mathcal{D}_{\text{test}}$}
    		\STATE Construct the baseline rationale $b_i$ 
    		\STATE $\text{\ourmetric}(x_i, y_i, r_i)$ \\$=-\log g'[b_i](y_i) + \log g[r_i, b_i](y_i)$
    		\STATE $\mathcal{S}.\text{add}(\text{\ourmetric}(x_i, y_i, r_i))$
    		\ENDFOR
    		\STATE $\ourmetric_\mathcal{D} = \text{mean}(\mathcal{S})$
    		\STATE {\bf Output}: $\mathcal{S}$, $\ourmetric_\mathcal{D}$
  \end{algorithmic}
\end{algorithm}

Algorithm \ref{alg:compute_rev} shows the process of computing both pointwise and aggregate \ourmetric scores. 
The higher the \ourmetric score, the more additional (\textit{new} and \textit{ relevant}) information the rationale $r$ contains to explain the label beyond the baseline rationale $b$. 
$\text{\ourmetric}(x_i, y_i, r_i)$ can take positive, negative, or zero values. 
When $\text{\ourmetric}(x_i, y_i, r_i)>0$, the rationale \textbf{supplies additional new information} for supporting the label (e.g., $r_1^*$ in Fig. \ref{fig:illustrations}); 
when $\text{\ourmetric}(x_i, y_i, r_i)=0$, the rationale \textbf{provides no additional information} beyond the baseline (e.g., ${\hat{r}}_{1,a}$ in Fig. \ref{fig:illustrations}); 
and when $\text{\ourmetric}(x_i, y_i, r_i)<0$, the rationale \textbf{does \textit{not} support the label} (e.g., ${\hat{r}}_{1,b}$ in Fig. \ref{fig:illustrations}). 
\ourmetric can assign a positive score to a rationale for an incorrect prediction as long as the rationale supports it and provides additional information beyond a vacuous baseline rationale (e.g., ${\hat{r}}_{2}$ in Fig. \ref{fig:illustrations}).
Thus, \ourmetric cannot be seen as a replacement for prediction accuracy, but rather as an orthogonal metric to interpret the usefulness of a generated rationale for the model decision. 
\section{Experimental Setup}
\label{sec:setup}

We outline our experimental setup by describing the reasoning tasks and datasets (\S\ref{sec:datasets}), followed by the task and evaluation models (\S\ref{sec:settings}), and the baseline metrics for comparison (\S\ref{sec:baseline_methods}). 
Additional details on the setup are provided in Appendix \ref{sec:exp_setup_supp}.

\subsection{Datasets}
\label{sec:datasets} 
We explore two reasoning tasks, namely CommonsenseQA (CQA) and Natural Language Inference (NLI) across four datasets, all containing human-annotated free-text rationales. 
For CQA task, we use ECQA \citep{aggarwal-etal-2021-explanations}, CoS-E (v1.11; \citealp{rajani2019explain}) and QuaRTz \citep{tafjord2019quartz}. 
For both ECQA and CoS-E, each commonsense question is paired with five candidate choices and the task is to select an answer from the candidates. 
ECQA contains higher quality human-written rationales compared to CoS-E \citep{aggarwal-etal-2021-explanations, sun2022investigating}. 
QuaRTz is for open-domain reasoning about textual qualitative relationships, and the task is to select an answer from two options to the question based on the textual qualitative knowledge (rationale). 
For the NLI task, we use the e-SNLI \citep{camburu2018snli} dataset containing explanations for SNLI \citep{bowman-etal-2015-large}, where the task is given a premise to predict if a hypothesis entails, contradicts or is neutral to it.
More details on the datasets are in Appendix \ref{sec:exp_setup_data}.

\subsection{Task and Evaluation Models}
\label{sec:settings}
\paragraph{Task models} 
We choose T5 Large \citep{raffel2020exploring} as the task model (finetuned on ground-truth labels and rationales) to produce generated rationale-label pairs under three settings:
\begin{itemize}
	\item \outputr : Given an input text and the ground-truth label, generate a rationale. 
	\item \outputyr: Given an input text, generate a label followed by a rationale. 
	Since T5 decodes tokens sequentially, each R is generated conditioned on the predicted Y. 
	\item \outputry: Given an input text, generate a rationale followed by a label. 
	Here, we compute a likelihood for each candidate Y conditioned on R, and then select the most probable candidate. 
	This operation can improve the model prediction accuracy, while weakening the consistency and relevance between the generated rationales and predicted labels. 
\end{itemize}
After training, we collect three types of rationale-label pairs by applying the three task models on the test set of each dataset. In addition to these three settings, we also evaluate ground-truth labels paired with crowd-sourced rationales (\gold).

\paragraph{Constructing a Baseline with Vacuous Rationales} 
Given an input $x$ and a label $y$ (ground-truth or model-generated), we construct a baseline rationale $b$ by declaratively combining $x$ and $y$ into a  sentence. 
For the CQA task, we adopt a T5-3B model fine-tuned on a set of (\textit{question}, \textit{answer}, \textit{declarative sentence}) tuples \citep{demszky2018transforming} following \citet{chen2021can}.\footnote{\url{https://github.com/jifan-chen/QA-Verification-Via-NLI}} 
For the NLI task, we first use a template to convert (\textit{premise}, \textit{hypothesis}, \textit{label}) tuple into a baseline rationale: ``\textit{premise} \texttt{implies} / \texttt{contradicts} / \texttt{is not related to} \textit{hypothesis}''. 
Then we paraphrase these templated, vacuous NLI rationales using a pre-trained model \footnote{\url{https://huggingface.co/humarin/chatgpt_paraphraser_on_T5_base}} in order to prevent the evaluators from learning the template patterns.
\autoref{tab:construct_rat} shows some examples of constructed vacuous baseline rationales.

\paragraph{Training Evaluation Models, $g$ and $g'$}
We train two evaluation models, $g$ and $g'$, which take $[r, b]$ and $b$ as inputs, respectively (see \autoref{eq:pcvi} in \S\ref{sec:method}).  
Both evaluators are based on fine-tuning T5 Large \citep{raffel2020exploring} models. 
We use the training set $\mathcal{D}_{train}=\{(x, y^*, r^*)\}$, where $\{y^*\}$ and $\{r^*\}$ are gold labels and human-annotated rationales, respectively. 
We construct baseline rationales $\{b^*\}$ based on $\{(x, y^*)\}$. 
The objective is to maximize the log-likelihood of $y^*$ given $[r^*, b^*]$ or $b^*$. 
After training, the evaluation models are applied to evaluate a rationale-label pair $(y, r)$ w.r.t. an input $x$. 
The rationale-label pair $(y, r)$ can be model-generated and the label may not be ground-truth (e.g., $y_2$ in Fig. \ref{fig:illustrations}), while \ourmetric is able to provide an assessment on the rationale along the two dimensions (\S\ref{sec:intro}). 
We refer readers to the Appendix \ref{sec:compare_evaluators} for results of using T5 Base, BART Large \citep{lewis2020bart}, and GPT-2 Large \citep{radford2019language} as evaluation model architectures.

\subsection{Other Metrics for Rationale Evaluation}
\label{sec:baseline_methods}
We compare with two existing automatic metrics for free-text rationale evaluation: LAS \citep{hase-etal-2020-leakage} and RQ \citep{wiegreffe-etal-2021-measuring}.
Analogous to our evaluation models, both approaches use proxy models; we use the same architecture (T5 Large) across metrics in our reported results.
\paragraph{Leakage-Adjusted Simulatability (LAS)}
\citet{hase-etal-2020-leakage} evaluate the quality of free-text rationales via a proxy model, trained with the task model outputs as labels and original input texts combined with rationales as input sequences. 
The metric computes the difference between its prediction accuracy on the predicted label when the rationale is included into the input vs. when it is not, $\mathbbm{1}[\hat{y} \mid x, \hat{r}] - \mathbbm{1}[\hat{y} \mid x]$, averaged over examples grouped based on whether they leak labels or not. 
The final LAS score is given by the macro average across groups.

\paragraph{Rationale Quality (RQ)}
\citet{wiegreffe-etal-2021-measuring} propose a variant of the simulatability in \citet{hase-etal-2020-leakage}. 
The main difference is that gold labels are used to train the model proxy and evaluate rationale quality. 
Specifically, the quality of a rationale $\hat{r}$ is measured as $\mathbbm{1}[y^* \mid x, \hat{r}] - \mathbbm{1}[y^* \mid x]$, where $y^*$ is the gold label. 
RQ is the average score over all test examples without considering label leakage. 

\begin{figure*}[ht]
  \centering
  \includegraphics[width=0.8\textwidth]{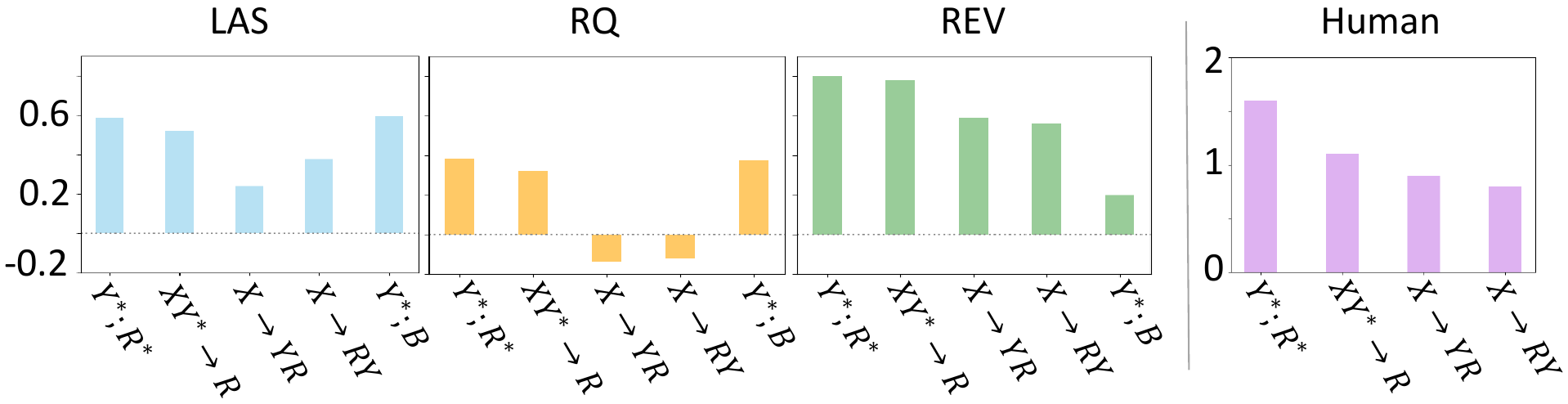}
  \caption{\label{fig:ecqa_eval} Left: Automatic evaluation results of LAS, RQ and \ourmetric for rationale-label pairs on the ECQA test set. Right: Human evaluation for rationale-label pairs on 230 randomly selected examples from the ECQA test set.
  }
\end{figure*}

\section{Evaluating \ourmetric}
\label{sec:exp}
We first compare \ourmetric with existing metrics (\S\ref{sec:compare_metrics}) and human judgments (\S\ref{sec:human_eval}) on the ECQA dataset, as well as show \ourmetric on other CQA and NLI benchmarks.
We then test the sensitivity of different metrics to input perturbations (\S\ref{sec:senitivity_test}). 
Next, we apply \ourmetric to generations via few-shot prompting  (\ref{sec:eval_prompting}). 
Additional experiments are listed in Appendix \ref{sec:exp_supp}.

\subsection{Comparison Between Evaluation Metrics}
\label{sec:compare_metrics}
We compare \ourmetric with LAS and RQ, in evaluating different rationale-label pairs on the ECQA dataset. 
In addition to \outputr, \outputyr, \outputry, and (\gold), we also explore the evaluation on the vacuous baseline rationales (\base) that are constructed with ground-truth labels. 
LAS, RQ and \ourmetric are not directly comparable due to different comparison scales and criteria (e.g., log-probability vs. accuracy); hence, our focus remains on the ranking over different sources of rationale-label pairs. 

Results are shown in Figure \ref{fig:ecqa_eval} (left panel). 
All three metrics rank the crowdsourced rationales (\gold) in ECQA the highest. 
While by definition, \ourmetric for vacuous rationales (\base) is low, both LAS and RQ scores for these rationales are quite high, showing that these metrics are incapable of measuring the amount of additional information in rationales. 
Intuitively, we expect weaker rationale-label consistency in \outputry setting compared to \outputyr, as the labels are forcefully selected among the candidates as opposed to being freely generated by the task model (\S\ref{sec:settings}). 
While \ourmetric is able to capture this intuition and ranks \outputyr higher than \outputry, LAS and RQ have a different ranking. 
Qualitative results comparing all three metrics are provided in Table \ref{tab:qualitative_analysis} in Appendix \ref{sec:qualitative_results}; Table \ref{tab:neg_rev} qualitatively analyzes rationales with negative \ourmetric scores.

We additionally analyze \ourmetric for ``input-irrelevant rationales'': sentences extracted from a knowledge base that contain the ground-truth labels but do not necessarily explain the labels w.r.t. the inputs.
Results in Appendix \ref{sec:label_relate_sents} show that \ourmetric penalizes such irrelevant rationales.

Next, we apply \ourmetric to evaluate crowdsourced and model generated rationale-label pairs (\gold, \outputr, \outputyr, \outputry) across different datasets. 
For each dataset, the evaluation models are trained on the training set with gold labels and crowdsourced rationales. 
The results are shown in \autoref{tab:cvi_datasets}. 
We observe that the gold rationales in the ECQA dataset achieve higher \ourmetric score than those in CoS-E. 
This observation is in line with the known quality issues of crowdsourced rationales in CoS-E \citep{aggarwal-etal-2021-explanations, sun2022investigating}.  
Interestingly, model-generated rationales (\outputr) have higher \ourmetric score than crowdsourced rationales for CoS-E (see examples in Table \ref{tab:cose_exp}). 
Please see Appendix \ref{sec:qualitative_cose} for a qualitative analysis on CoS-E rationales.
QuaRTz has better quality of rationales compared to ECQA, CoS-E, and e-SNLI. 
In the case of e-SNLI, the problem is severe as most of the crowdsourced or generated rationales do not provide reasoning but rather follow a label-specific template e.g., \textit{A implies (that) B} \citep{kumar2020nile, brahman2021learning}. 
\begin{table}[t] 
 	\centering
  \scalebox{0.85}{
 	\begin{tabular}{ccccc}
 		\toprule
 		\multirow{2}{*}{Datasets} & \multicolumn{4}{c}{Rationale-label pairs} \\
		\cmidrule(lr){2-5}
		 &  \gold & \outputr & \outputyr & \outputry  \\
 		\midrule
 		ECQA & 0.7943 & 0.7806 & 0.5840 & 0.5599 \\
 		CoS-E & 0.2415 & 0.4050 & 0.2308 & 0.1198 \\
 		QuaRTz & 1.3919 & 1.3696 & 1.3449 & 1.0170 \\
 		\midrule
 		e-SNLI & 0.0752 & 0.0079 & 0.0055 & 0.0047 \\
 		\bottomrule
 	\end{tabular}
  }
 	\caption{\ourmetric scores of different types of rationale-label pairs on the four datasets.}
 	\label{tab:cvi_datasets}
\end{table}

\begin{figure*}[ht]
	\centering
	\subfigure[\outputry, LAS]{
		\label{fig:RY_LAS}
		\includegraphics[width=0.32\textwidth]{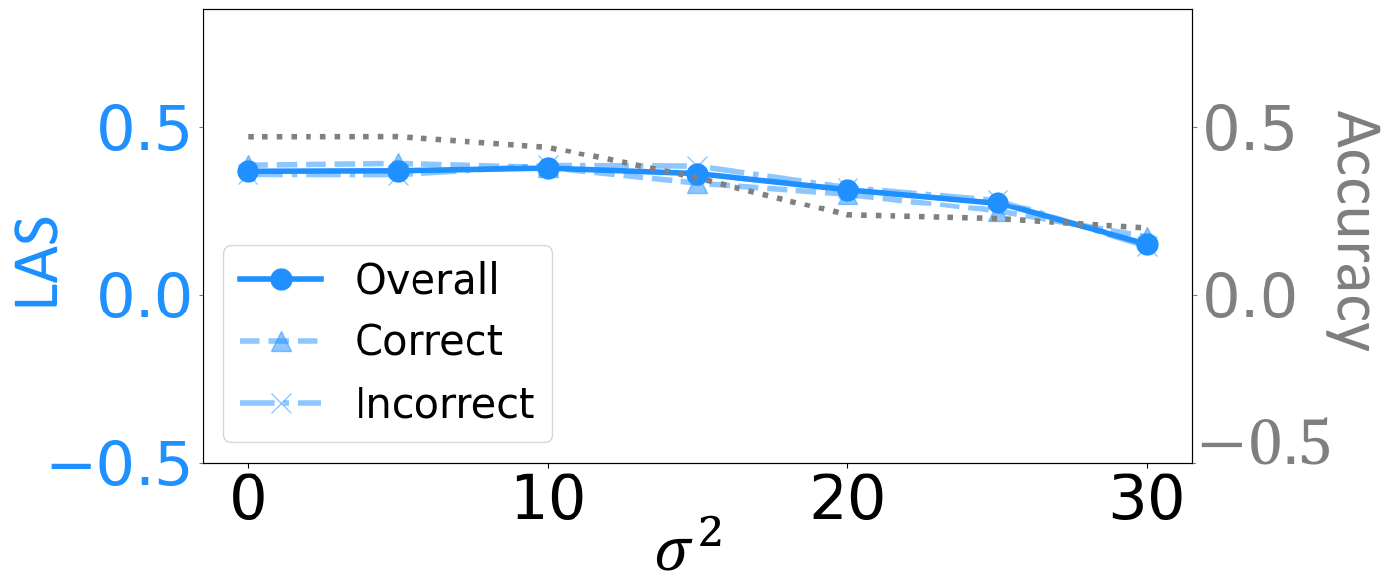}}
	\subfigure[\outputry, RQ]{
		\label{fig:RY_RQ}
		\includegraphics[width=0.32\textwidth]{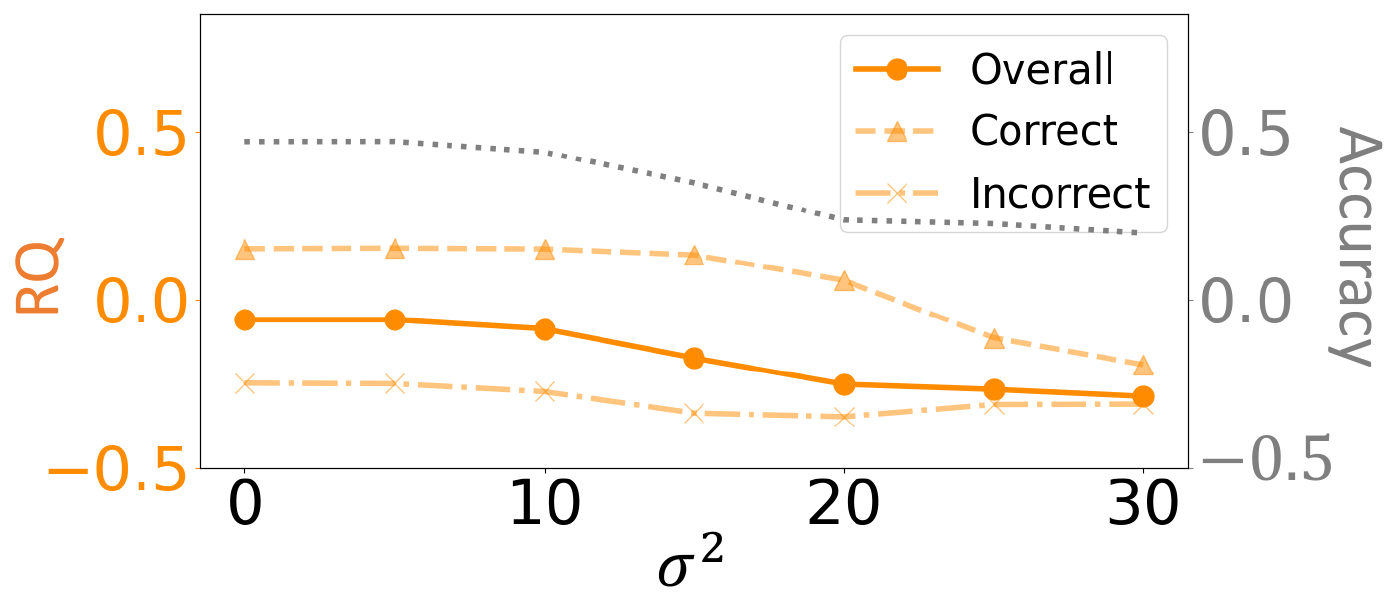}}
	\subfigure[\outputry, \ourmetric]{
		\label{fig:RY_CVI}
		\includegraphics[width=0.32\textwidth]{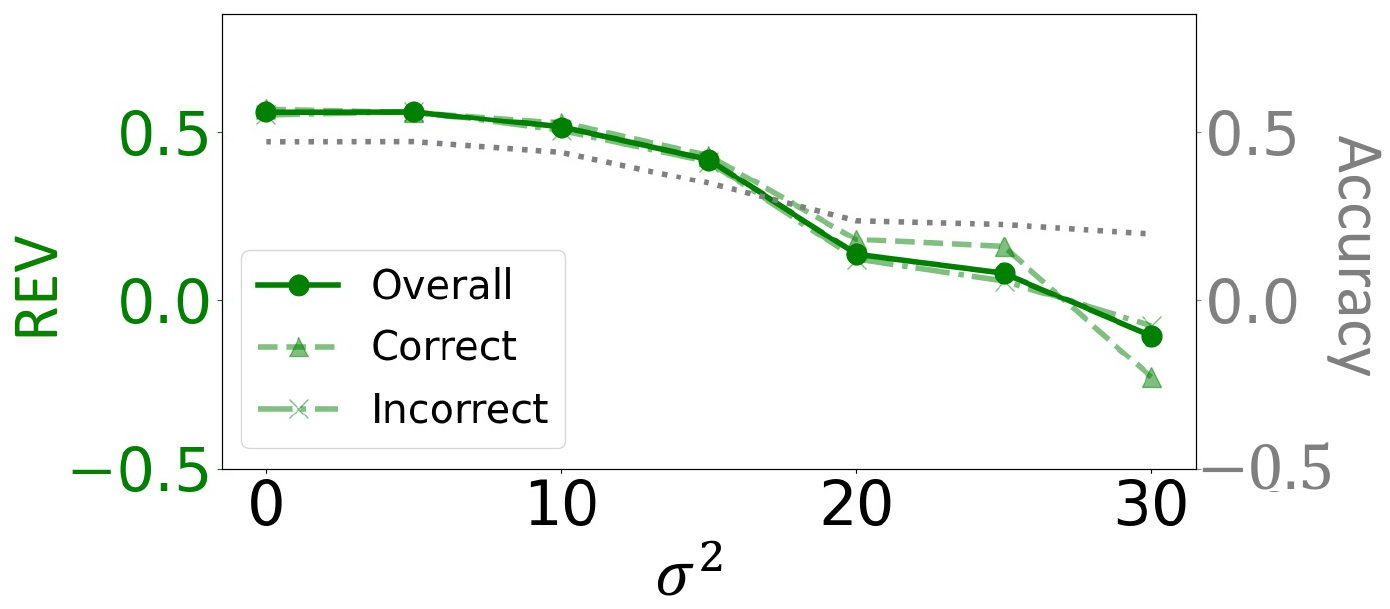}}
	\subfigure[\outputyr, LAS]{
		\label{fig:YR_LAS}
		\includegraphics[width=0.32\textwidth]{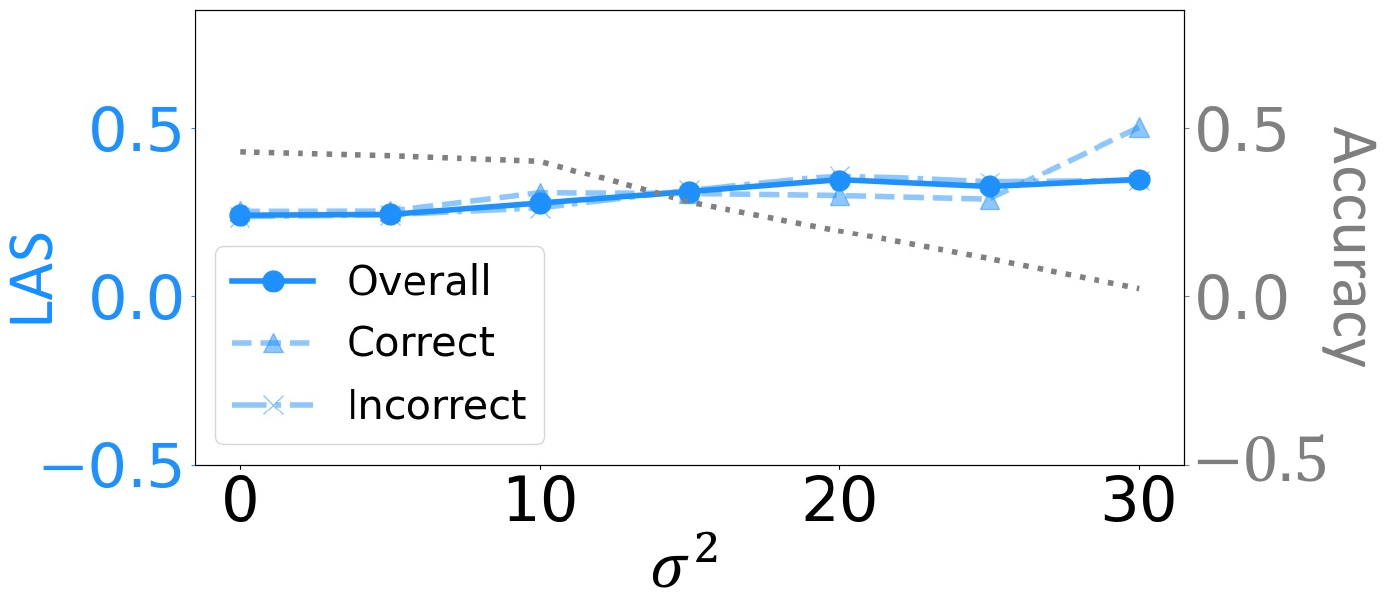}}
	\subfigure[\outputyr, RQ]{
		\label{fig:YR_RQ}
		\includegraphics[width=0.32\textwidth]{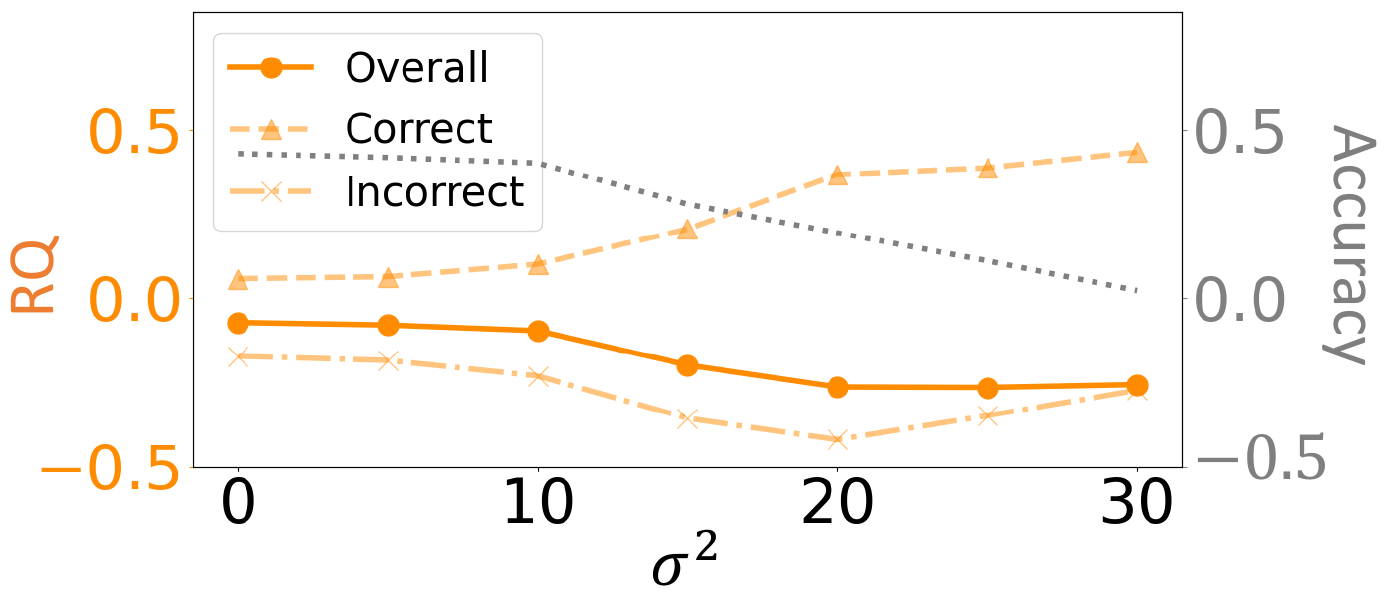}}
	\subfigure[\outputyr, \ourmetric]{
		\label{fig:YR_CVI}
		\includegraphics[width=0.31\textwidth]{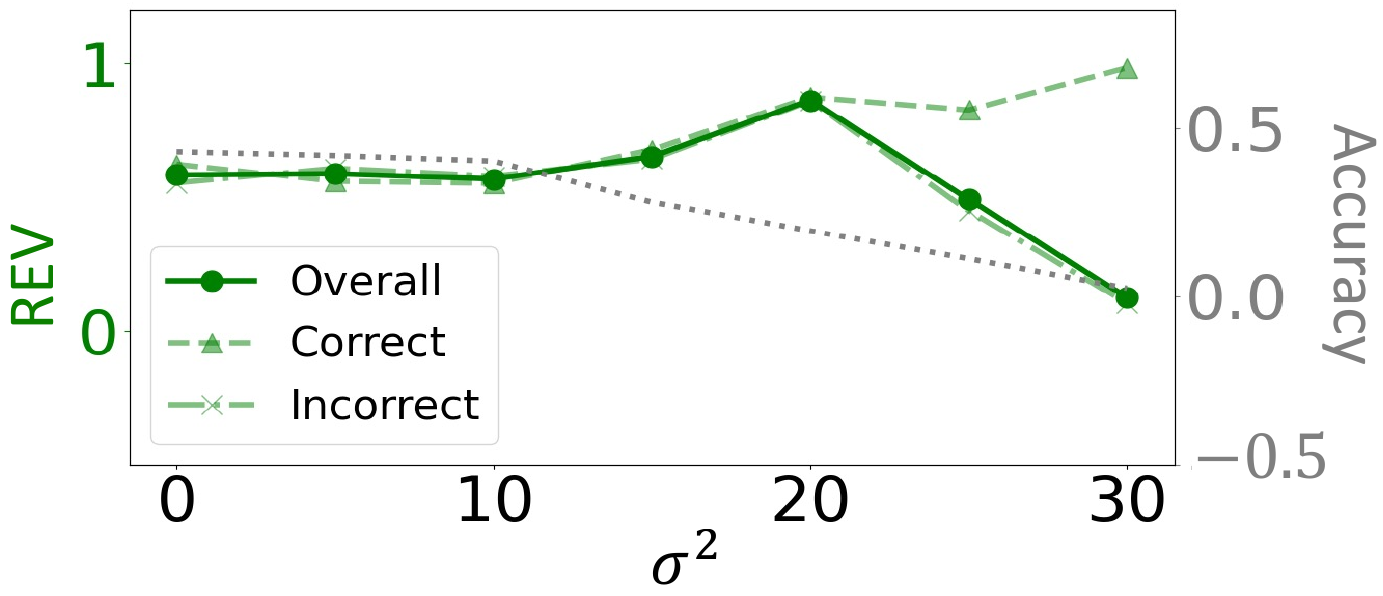}}
	\caption{
	Sensitivity test results of \ourmetric, LAS and RQ for \outputry and \outputyr on the ECQA dataset. The $X$-axis shows different levels of noise ($\sigma^{2}$). 
	We plot the curve of Accuracy (model prediction accuracy) vs. Noise in gray dashed line. 
	We also separate the evaluation results on populations on which the model predictions are correct (``Correct'') or incorrect (``Incorrect'') in addition to the overall evaluation on all test examples (``Overall'').
	}
	\label{fig:sensitivity_test}
\end{figure*}

\subsection{Human Evaluation}
\label{sec:human_eval}

We collect crowdworker judgments via Amazon Mechanical Turk to understand how \ourmetric correlates with human judgments of rationales. 
We randomly sample 230 examples from the ECQA test set and ask workers to evaluate the four types of rationale-label pairs (\gold, \outputr, \outputyr, \outputry) for each example.\footnote{
We do not consider (\base) because we have trained workers to recognize baseline rationales as vacuous.} 
We present workers with a question (input text), an answer (label) and an explanation (rationale), and ask them whether the explanation justifies the answer (\textit{yes/no}). 
If they answer \textit{yes}, we further ask them to evaluate the amount of additional information supplied by the explanation that explains \textit{why} the answer might have been chosen for the question by choosing from \textit{none / little / some / enough}, corresponding to a 4-point Likert-scale (0/1/2/3). 
We collect 3 annotations per instance and use majority vote to decide whether the rationale can justify the label. 
If \textit{yes}, we take the average over the 3 human-annotated scores as the amount of information. 
Otherwise, we give a score of -1. 
More details of human evaluation are in Appendix \ref{sec:human_eval_supp}.

Results are shown in the right panel of Fig. \ref{fig:ecqa_eval}, where the ranking of the four types of rationale-label pairs is \gold $>$ \outputr $>$ \outputyr $>$ \outputry. 
While LAS and RQ rank \outputry better than \outputyr (see the left part of Fig.~\ref{fig:ecqa_eval}), the ranking from \ourmetric is more consistent with human judgments, suggesting its effectiveness in evaluating rationales.

\subsection{Is \ourmetric sensitive to input perturbations?}
\label{sec:senitivity_test}

A robust metric should be sensitive to the change of rationale-label pairs and reflect their relationships under input perturbations.
We test the sensitivity of all automatic metrics to input ($X$) perturbations in the task model, under two settings: \outputyr and \outputry.
Following \citet{wiegreffe-etal-2021-measuring}, we add zero-mean Gaussian noise $\mathcal{N}(0, \sigma^{2})$ to input word embeddings during inference, inducing task models to produce progressively degenerate rationales and labels. 
Results in Fig. \ref{sec:senitivity_test} indicate that \ourmetric (b) and RQ (c) follow similar trends as for \outputry. 
However, LAS is less sensitive to noise for both joint models, \outputry (a) and \outputyr (d). 
Since the proxy model for LAS was trained on the task models' predicted labels and generated rationales, it can overfit to the degenerate rationale-label pairs under input perturbations, hence being less sensitive to input noise during inference.
The largest differences between \ourmetric and RQ are for \outputyr.  
We observe the task model can predict incorrect labels and then make up reasonable-sounding rationales for its wrong predictions under certain input perturbations; prior work also reports this finding \citep{narang2020wt5,wiegreffe-etal-2021-measuring}. 
\ourmetric does not drop under a certain amount of input perturbations (e.g., $\sigma^{2} \leq 20$) in Fig. \ref{fig:sensitivity_test} (f), likely because the generated rationales still provide new information for describing both correct and incorrect labels (also see the example in Table \ref{tab:sensitivity_test}).
However, as the noise exceeds the certain level, \ourmetric decreases indicating that the task model is no longer able to make up rationales for very noisy inputs. 
On the other hand, the behavior of RQ in Fig. \ref{fig:sensitivity_test} (e) is quite different to \ourmetric. 
Since RQ is computed based on gold labels (\S\ref{sec:baseline_methods}), it has reduced sensitivity to input perturbations. 
When the prediction accuracy decreases, the overall evaluation of RQ is dominated by the results on incorrect predictions, as shown in Fig. \ref{fig:sensitivity_test} (e). 
We refer readers to the Table \ref{tab:sensitivity_test} in Appendix \ref{sec:sensitivity_test_supp} for qualitative analysis on sensitivity test.

\subsection{Evaluating Rationales in Few-shot Prompting}
\label{sec:eval_prompting}

We test the ability of \ourmetric in evaluating rationales generated by few-shot prompting, and get insights into the reasoning and prediction processes of large language models (e.g., GPT-3). 

\begin{figure}[ht]
\centering
  \includegraphics[width=0.95\linewidth]{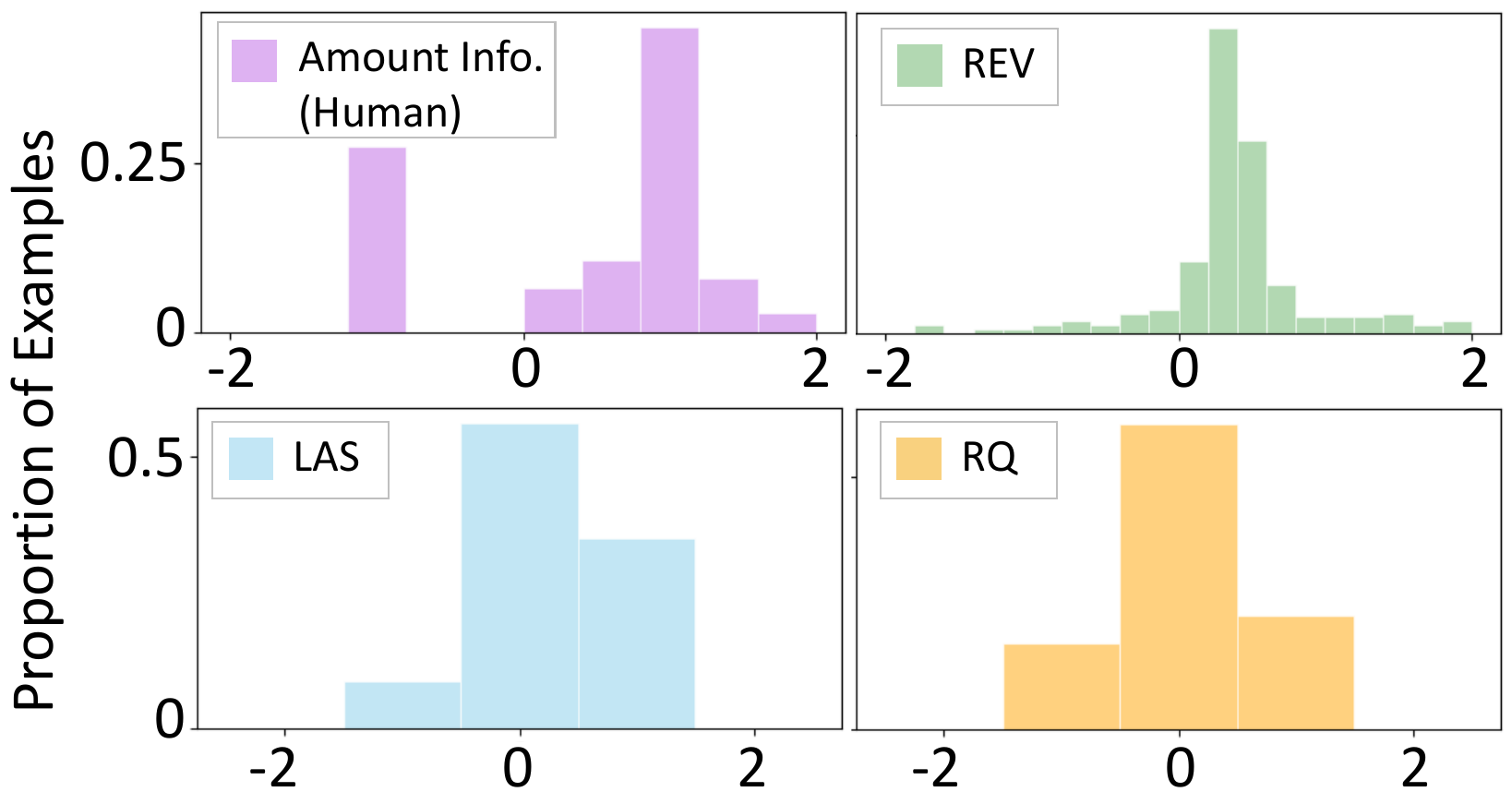}
  \captionof{figure}{Histograms of human-annotated amount of information and pointwise \ourmetric, LAS and RQ scores on GPT-3 few-shot prompted rationales for gold labels. 
  }
  \label{fig:eval_gpt3_rat}
\end{figure}

\paragraph{GPT-3 Rationales for Gold Labels} \citet{wiegreffe2021reframing} collected 250 high quality free-text rationales generated by few-shot prompting with GPT-3 \citep{brown2020language} for CQA (given gold labels). 
Each example was assessed by 3 crowdworkers. 
We focus on two aspects of their annotations: ``supports the gold label'' and ``amount of information''. 
Crowdworkers provide a \textit{yes / no} answer to justify whether a rationale supports the corresponding gold label. 
Only when the answer is \textit{yes}, they are further asked to evaluate the amount of information contained in the rationale for justifying the label. 
The amount of information is roughly categorized into 3 levels: ``Not Enough'', ``Enough'', ``Too Much'', each annotated with a Likert-scale score.\footnote{The original human-annotated scores w.r.t. the three levels are: -1, 0, 1. Since \citet{wiegreffe2021reframing} suggest ``a value of 0 is preferred to a value of 1'', we map the scores $\{$-1, 0, 1$\}$ to $\{$0, 1, 2$\}$ accordingly. The value ``-1'' is then given to examples annotated as ``not supporting gold labels''.} 
In Fig. \ref{fig:eval_gpt3_rat}, we compare human annotation scores for amount of information\footnote{We take majority vote to decide ``supports the gold label", and average ``amount of information" over 3 workers.} with the pointwise scores obtained by three automatic metrics, LAS, RQ, and \ourmetric. 
For automatic metrics, the evaluation models of \ourmetric and the proxy models of LAS and RQ are trained on the ECQA training set with gold labels and human-annotated rationales (\S\ref{sec:settings}). 
We observe that \ourmetric provides finer-grained assessment of the information contained in rationales compared to LAS and RQ which only take $\{$-1, 0, 1$\}$ values. 
When LAS and RQ are zero, it is unclear whether the rationale supports the label or not because the model proxy may predict the label based on the input only. 
The judgments of \ourmetric on whether rationales support labels (\ourmetric $> 0$ ) are close to human judgments (i.e., $80\%$ agreement). 
The support rates of LAS and RQ are relatively low, i.e. $35\%$ and $23\%$, while a large portion ($56\%$ and $60\%$ respectively) corresponds to a zero LAS / RQ score.

\begin{figure}[ht]
\centering
  \includegraphics[width=0.8\linewidth]{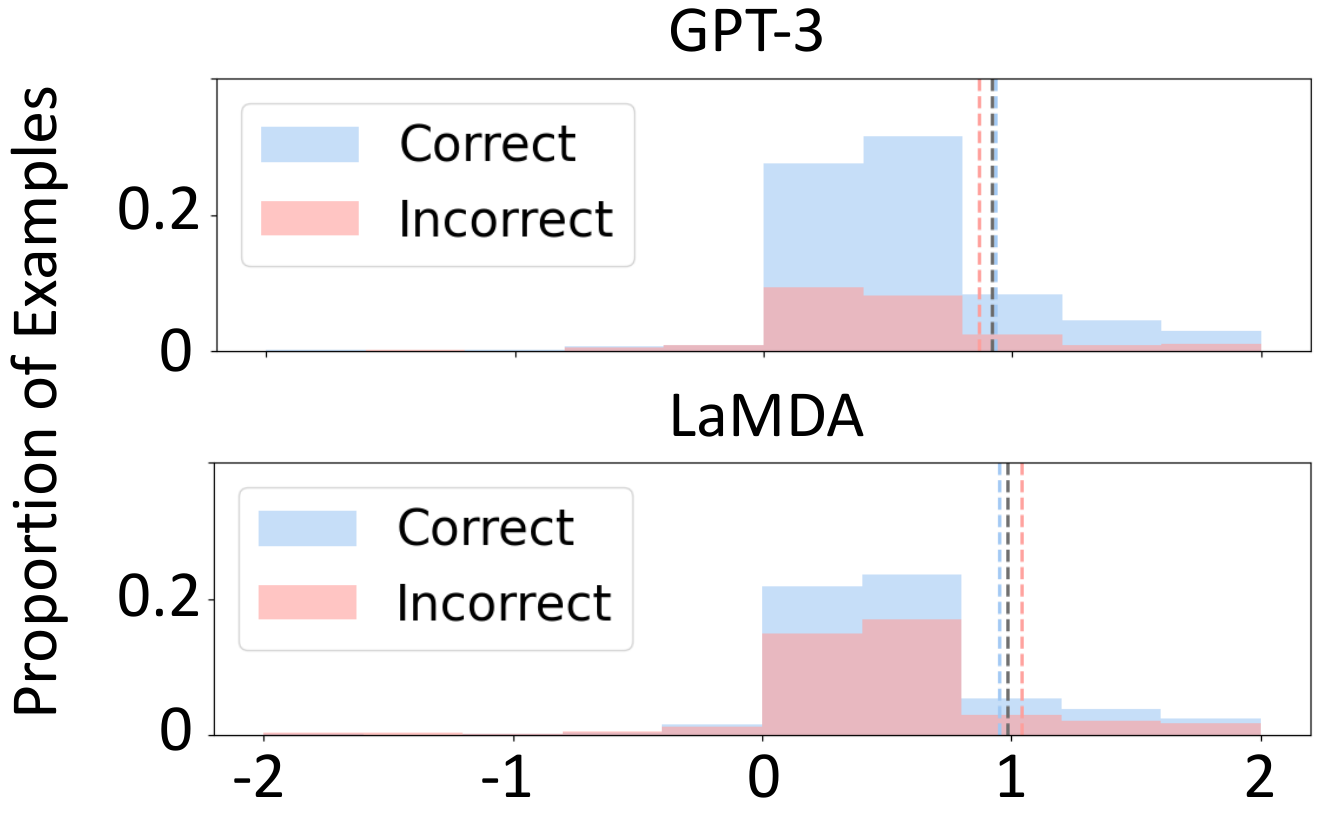}
  \captionof{figure}{Distributions of \ourmetric for rationales w.r.t. correct and incorrect predictions produced by GPT-3 and LaMDA respectively. 
  The average \ourmetric scores over all instances, correctly predicted instances and incorrectly predicted instances are marked by gray, blue and red dashed lines respectively.}
  \label{fig:cot_test}
\end{figure}

\paragraph{Chain of Thought Rationales} \citet{wei2022chain} propose \textit{chain of thought prompting} to teach large language models to produce intermediate reasoning steps (rationales) before prediction, which improves their prediction performance on a range of reasoning tasks (e.g., arithmetic and symbolic reasoning). 
However, the reported improvement is trivial for CQA \citep{wei2022chain}, which motivates us to evaluate the intermediate rationales w.r.t. model predictions. 
We apply \ourmetric to analyze the generated rationales during intermediate reasoning steps and final predicted labels from GPT-3 text-davinci-002 \citep{brown2020language} and LaMDA 137B \citep{thoppilan2022lamda}.\footnote{Available at \url{https://github.com/jasonwei20/chain-of-thought-prompting}}

Figure \ref{fig:cot_test} shows the distributions of \ourmetric for correctly and incorrectly predicted instances from GPT-3 and LaMDA, respectively. 
For both GPT-3 and LaMDA, the \ourmetric distributions of correct and incorrect predictions are similar and most instances have positive \ourmetric scores. 
The average \ourmetric scores over correct and incorrect predictions (blue and red dashed lines, resp.) are close, especially for GPT-3. 
This is consistent with our observation that most generated rationales from the two models are describing their predicted labels. 
The prediction accuracy of GPT-3 is much higher than that of LaMDA ($77\%$ vs. $59\%$), while the average \ourmetric scores over all instances (gray dashed lines) are close (0.92 vs. 0.99). 
An insight we obtain is that the generated intermediate reasoning steps (rationales) support models' predictions (consistent \ourmetric scores), but cannot guarantee their correctness (discrepant accuracies between GPT-3 and LaMDA).
This partially explains the minor improvement of chain of thought prompting on CQA. 
\section{Related Work}
\label{sec:relate}
Model rationales broadly fall into two categories: extractive rationales and free-text rationales. 
Extractive rationales contain some important features extracted from input texts that make models produce final predictions \citep{lei2016rationalizing, deyoung2020eraser, jain2020learning, schulz2019restricting}. 
Free-text rationales are produced by generative models in the form of natural language. 
Compared to extractive rationales, free-text rationales explain model predictions in a more human-like way and fill the gap in explaining reasoning tasks \citep{camburu2018snli, narang2020wt5, rajani2019explain, kumar2020nile, brahman2021learning}.

Evaluations on extractive rationales have been well studied, generally from two perspectives — faithfulness and plausibility \citep{deyoung2020eraser, pruthi2022evaluating, chan2022unirex}. 
Faithfulness measures to which extent rationales reflect the true reasoning process of models, while plausibility evaluates how convincing rationales are to humans \citep{jacovi2020towards}. 
Other perspectives include the ability of rationales in helping a student model simulate a teacher model \citep{pruthi2022evaluating} or bridging the communication between a classifier and a layperson \citep{treviso2020explanation}. 
Existing automatic metrics for free-text rationales focus on rationale-label association, and measure the utility of a rationale based on how much it helps a model proxy predict the given label (inspired by human simulatability \cite{doshi2017towards}) \citep{hase-etal-2020-leakage} or the gold label \citep{wiegreffe-etal-2021-measuring} given the input. 
\citet{chan2022frame} further propose a framework to evaluate the automatic metrics. 
However, none of them consider measuring the amount of additional new information in free-text rationales. 
\citet{sun2022investigating} conduct a human study on the additional knowledge provided by free-text rationales. 
This work is the first that proposes an automatic metric to quantify the new information in free-text rationales.
\section{Conclusion}
\label{sec:con}
We introduce \ourmetric, an information-theoretic measure to evaluate the amount of new, label-relevant information in free-text rationales, \textit{beyond} the information contained in the input.
We empirically demonstrate the advantage of \ourmetric compared to existing metrics focusing simply on label-rationale association, and show that \ourmetric is more consistent with human judgments.
\ourmetric also offers insights into evaluating rationales generated via few-shot prompting. 
While we recommend the usage of \ourmetric alongside traditional performance metrics, future work might explore a combined metric to measure the correctness of a prediction as well as the informativeness of the rationale towards this prediction.
Ultimately, free-text rationales are for the benefit of human users and there exist multiple criteria for human utility of rationales \cite{joshi2023are}, beyond label relevance and informativeness.

\section*{Limitations} 
In its current formulation, \ourmetric might reward a rationale for an incorrect prediction as long as the rationale supports the prediction with relevant additional information. 
Additionally, our metric does not consider the factuality of rationales.
Future work might explore evaluation that penalizes rationales which support incorrect predictions, thus bridging together predictive performance with interpretability metrics. 
We considered a single declarative construction for baseline rationales and leave analyzing how different baseline construction impacts our metric to future work.
Another limitation is that the utility of \ourmetric depends on the quality of crowd-sourced rationales used to train the evaluator. 
Building a good automatic metric \ourmetric requires high-quality rationales that provide sufficient new information (e.g., commonsense knowledge) to explain the corresponding labels. 
The architecture of evaluation models also has an impact on \ourmetric evaluation. 
Using different evaluator architectures may result in varying \ourmetric scores, as discussed in Appendix \ref{sec:compare_evaluators}.

\section*{Ethics Statement}
All datasets used in this work are public, and deal with situations encountered in daily life; these are the examples provided for human annotation.
Generated rationales sometimes contain non-factual statements or misinformation.
While it is plausible that some rationales generated by the model or some data instances might contain offensive material, to the best of our knowledge we did not encounter such examples. 
We did not collect any personal information (e.g. demographics and identities) of participants in any of the human evaluation experiments.

\section*{Acknowledgements}
We thank the anonymous reviewers for many valuable comments. 
We thank Sarah Wiegreffe, Aaron Chan, and the Mosaic team at the Allen Institute for AI for helpful discussions and suggestions.

\bibliography{anthology,custom}
\bibliographystyle{acl_natbib}

\clearpage
\appendix
\onecolumn
\section{Properties of Conditional $\mathcal{V}$-information}
\label{sec:property_cvi}

As proved by \citet{hewitt-etal-2021-conditional}, CVI has several useful properties:
\begin{enumerate}
    \item \textit{Non-Negativity}: $I_{\mathcal{V}}(R \rightarrow Y \mid B) \geq 0$.
    \item \textit{Independence}: If $Y$ and $B$ are jointly independent of $R$, then $I_{\mathcal{V}}(R \rightarrow Y \mid B)=0$.
    \item \textit{Monotonicity}: If $\mathcal{U} \subseteq \mathcal{V}$, then $H_{\mathcal{V}}(Y \mid B) \leq H_{\mathcal{U}}(Y \mid B)$.
\end{enumerate}
An implication from \textit{Monotonicity} is complex models (e.g., pre-trained language models) might do better than simpler ones (e.g., linear models) in estimating $\mathcal{V}$-usable information. 
Since CVI measures the additional $\mathcal{V}$-usable information in $R$ about $Y$ beyond what's already extracted from $B$ by models in $\mathcal{V}$, it grounds the goal of the proposed metric \ourmetric.

\section{Additional Details on the Experimental Setup}
\label{sec:exp_setup_supp}
\subsection{Datasets}
\label{sec:exp_setup_data}
For CQA task, we use ECQA \citep{aggarwal-etal-2021-explanations}, CoS-E (v1.11) \footnote{We use the version v1.11 where each question is paired with 5 answer choices, for comparison with ECQA.} \citep{rajani2019explain} and QuaRTz \citep{tafjord2019quartz}. 
Both ECQA and CoS-E originate from the CommonsenseQA dataset \citep{talmor2018commonsenseqa}, where each commonsense question is paired with 5 candidate choices and the task is to select an answer from the candidates. 
ECQA contains higher quality free-text rationales compared to CoS-E, in terms of comprehensiveness, coherence, non-redundancy, etc. \citep{aggarwal-etal-2021-explanations, sun2022investigating}. 
QuaRTz is an open-domain reasoning task about textual qualitative relationships. 
Each instance contains a situated qualitative question, two answer options and a knowledge statement. 
The task is to select an answer from the two options to the question based on the textual qualitative knowledge. 
We use the knowledge statement as a free-text rationale since it explains why the answer is to the question. 
For NLI task, we use e-SNLI \citep{camburu2018snli} which is an extension of SNLI \citep{bowman-etal-2015-large} with augmented free-text human-written rationales. 
The task is to predict the entailment relationship between a premise and a hypothesis. 
\autoref{tab:datasets} shows the summary statistics of the four datasets.\footnote{Since CoS-E does not provide rationales for instances in the test set, we use the original development set as the test set and hold out 10\% of training data as the new development set. 
For e-SNLI, we follow \citet{hase-etal-2020-leakage} and randomly sample 10\% of training data to form the training set for finetuning our models.}
\begin{wrapfigure}{R}{0.35\textwidth}
 	\centering
 	\begin{tabular}{cccccc}
 		\toprule
 		Datasets  & \textit{\#train} & \textit{\#dev} & \textit{\#test} \\
 		\midrule
 		ECQA & 7598 & 1090 & 2194 \\
 		CoS-E & 8766 & 975 & 1221 \\
 		QuaRTz & 2696 & 384 & 784 \\
 		e-SNLI & 54933 & 9842 & 9824 \\
 		\bottomrule
 	\end{tabular}
 	\caption{Summary statistics of the datasets, where \textit{\#} counts the number of examples in the \textit{train/dev/test} sets.}
 	\label{tab:datasets}
\end{wrapfigure}

\subsection{Models}
\label{sec:exp_setup_model}
We use Huggingface Transformers \citep{wolf2020transformers} to access all task and evaluation models. 
We train each model for up to 20 epochs with a learning rate $5e-6$ and a batch size $8$. 
All experiments were performed on a single NVIDIA RTX 8000 GPU. 
Table \ref{tab:in_out_format} shows input-output formattings of different task models for different tasks.
\begin{table*}[h]
\small
	\centering
	\begin{tabular}{P{1cm}P{8cm}P{4.5cm}}
		\toprule
		Type & Input & Output \\
		\midrule
		\multirow{2}{*}{\outputr} & CQA: [question] {question} [choice] {choice-1} ... [choice] {choice-n} [answer] gold label [rationale] 
		& \multirow{2}{*}{rationale  \textless eos\textgreater}  \\
		& NLI: [premise] premise [hypothesis] hypothesis [answer] gold label [rationale] & \\
		\midrule
        \multirow{2}{*}{\outputyr} & CQA: [question] {question} [choice] {choice-1} ... [choice] {choice-n} [answer]
		& 
		\multirow{2}{*}{label [rationale] rationale  \textless eos\textgreater}  \\
		& NLI: [premise] premise [hypothesis] hypothesis [answer] & \\
		\midrule
		\multirow{2}{*}{\outputry} &
		CQA: [question] {question} [choice] {choice-1} ... [choice] {choice-n} [rationale]
		&
		\multirow{2}{*}{rationale [answer] label \textless eos\textgreater}  \\
		& NLI: [premise] premise [hypothesis] hypothesis [rationale] & \\
		\bottomrule
	\end{tabular}
	\caption{The input-output formatting of different task models.}
	\label{tab:in_out_format}
\end{table*}

\subsection{Comparison Between Evaluator Architectures}
\label{sec:compare_evaluators}
\begin{wrapfigure}{r}{0.5\textwidth}
  \centering
  \includegraphics[width=0.4\textwidth]{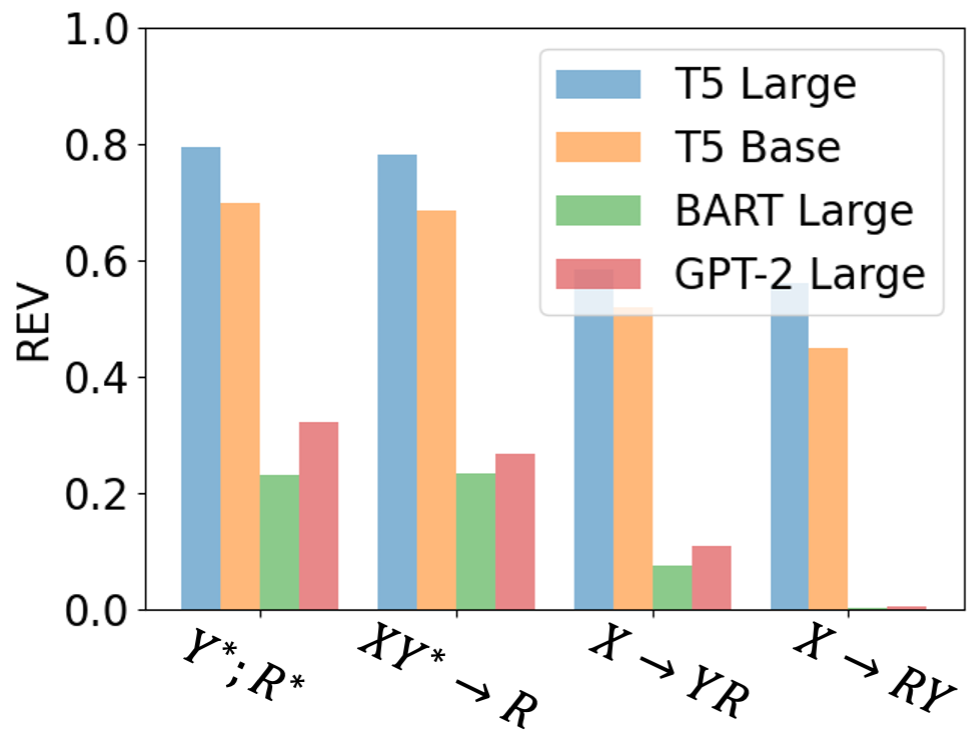}
  \captionof{figure}{\ourmetric for evaluating rationale-label pairs on the ECQA dataset with different evaluator architectures.}
  \label{fig:ecqa_model_cvi}
\end{wrapfigure}

We apply \ourmetric to evaluate different types of free-text rationales w.r.t. labels on the ECQA dataset. 
\autoref{fig:ecqa_model_cvi} shows \ourmetric scores of the four types of rationale-label pairs evaluated by four evaluator architectures. 
The ranking of the four groups of rationale-label pairs is consistent across the four evaluators, i.e. \gold $>$ \outputr $>$ \outputyr $>$ \outputry. 
This ranking is also consistent with human evaluation in  \S\ref{sec:human_eval}. 
Since ECQA contains high-quality crowdsourced rationales \citep{aggarwal-etal-2021-explanations}, it is expected that the \ourmetric of gold rationale-label pairs (\gold) is the highest. 
The \ourmetric of \outputr is close to that of \gold, indicating the task model (T5 Large) can produce good quality rationales when it is prompted with ground-truth labels. 
All four evaluators agree that the generated rationales of \outputyr contain more additional background information for explaining the predicted labels than those of \outputry. 
This is consistent with our design of the \outputry in \S\ref{sec:baseline_methods}, where the generated rationales and labels have weakened relevance. 
For each type of rationale-label pairs, the four evaluators capture different amount of conditional $\mathcal{V}$-information, while T5 Large consistently outperforms other three models.  
In the reported experiments \S\ref{sec:exp}, we use T5 Large as the evaluation model.

\section{Additional Experiments}
\label{sec:exp_supp}

\subsection{Qualitative Analysis of Different Metrics on ECQA}
\label{sec:qualitative_results}
Table \ref{tab:qualitative_analysis} shows the qualitative analysis of different metrics on the four types of rationale-label pairs (\gold, \outputr, \outputyr, \outputry) on the ECQA dataset. 
\ourmetric provides more accurate evaluations on those examples than LAS and RQ. 
\begin{table*}[h]
\small
	\centering
	\begin{tabular}{P{1cm}P{4cm}P{1.5cm}P{4cm}P{0.8cm}P{0.8cm}P{0.8cm}}
		\toprule
		\multirow{2}{*}{\footnotesize Type} & \multirow{2}{*}{\footnotesize Question} & \multirow{2}{*}{\footnotesize Label} & \multirow{2}{*}{\footnotesize Rationale} & \multicolumn{3}{c}{\footnotesize Metric} \\
		\cmidrule(lr){5-7}
		& & & & \footnotesize \ourmetric & \footnotesize LAS & \footnotesize RQ \\
		\midrule
		\multirow{3}{*}{\gold} & If you have a ticket and you are planning to eat hot dogs, where would you go? & baseball stadium & Hot dogs can be eaten at baseball stadium. When you go to a baseball stadium, you have a ticket and you may plan to eat hot dogs. & 0.32 & 0 & 0 \\
		\rule{0pt}{3ex}
		& How does a person go to space? & space shuttle & People go to space by a vehicle specially designed to travel to space. That vehicle is called a space shuttle. & 0.32 & 0 & 0 \\
		\rule{0pt}{3ex}
		& What is a dangerous outdoor activity for children? & sun themselves & Sunning themselves is a dangerous activity Children should not sun themselves & 0.28 & 1 & 1 \\
		\midrule
		\multirow{3}{*}{\outputr} & Where are old pictures kept? & attic & Attic is a place where old pictures are kept. & 0.34 & 1 & 0 \\
		\rule{0pt}{3ex}
		& What would you be if you comfort friend? & friendly & Comforting friend is a good thing. & 0.43 & 0 & 1 \\
		\rule{0pt}{3ex}
		& What do customers do to a waiter after the waiter serves customers? & pay to & Paying to a waiter is the action of paying. Waiters get paid to serve customers. & 0.10 & -1 & 0 \\
		\midrule
		\multirow{3}{*}{\outputyr} & Where is there likely to be more than one desk drawer? & \textcolor{red}{desk} & Desk drawer is a drawer used for storing office supplies. There is likely to be more than one desk drawer in office. & -4.75 & -1 & 1 \\
		\rule{0pt}{3ex}
		& What leads to someone's death when they are very depressed? & suicide & Suicide is the act of committing suicide. When someone is very depressed, suicide leads to their death. & 0.57 & 0 & 0 \\
		\rule{0pt}{3ex}
		& Where are you normally when you take a bath? & \textcolor{red}{hotel room} & Hotel room is a place where people stay. Bathing is normally done in hotel rooms. & 0.32 & 0 & -1 \\
		\midrule
		\multirow{3}{*}{\outputry} & What is likely heard by those going to a party? & laughter & People go to a party to meet new people. People are likely to hear laughter at the party. & -0.10 & 1 & 0 \\
		\rule{0pt}{3ex}
		& What would you do if you have excitement and do not want to stay in your house? & \textcolor{red}{go to gym} & Go to gym is to go to a place where you can express information. If you have excitement and do not want to stay in your house, then you would go somewhere. & 0.53 & 1 & 0 \\
		\rule{0pt}{3ex}
		& If you're caught committing murder, an injection can lead to your own what? & \textcolor{red}{die} & An injection can lead to one's own death. If you're caught committing murder, you can be injected into your own body and die. & 1.46 & 0 & 0 \\
	\bottomrule
	\end{tabular}
	\caption{Pointwise evaluation of \ourmetric, LAS and RQ on different types of rationale-label pairs. Incorrect labels are colored red.}
	\label{tab:qualitative_analysis}
\end{table*}

\subsubsection{Qualitative Analysis of Negative \ourmetric Scores in ECQA}
\label{sec:qualitative_neg_rev}
Table \ref{tab:neg_rev} shows some examples of \outputry with negative \ourmetric scores on the ECQA dataset. 
When \ourmetric $ < 0$, we observe in most cases the rationale does not support the given label, while indicating other labels, or something even beyond the label candidates (e.g., ``helicopter" in the second example), or they could repeat the input (e.g., the first example). 
The same observation holds for other types of rationale-label pairs.

\subsection{Additional Analysis on Label-Related But Input-Irrelevant ``Rationales''}
\label{sec:label_relate_sents}
\begin{table*}[h]
\footnotesize
	\centering
 \resizebox{\textwidth}{!}{ 
	\begin{tabular}{P{4cm}P{2cm}P{4cm}P{0.8cm}P{4cm}P{0.8cm}}
		\toprule
		Input & Label & Crowdsourced Rationale & \ourmetric & Input-Irrelevant GenericsKB Sentence & \ourmetric \\
		\midrule
            What form of government is most associated with kingdoms? & monarchy & Monarchy is a form of government with the monarch at the head. Monarchy is a form of government mostly associated with kingdoms. & 0.65 & Monarchies are countries. & -0.94\\
		\rule{0pt}{5ex}
		Bailey liked playing games against other people. He found it exhilarating.  What might Bailey like about games? & competitiveness & When a game is played against someone, it is a competition and it promotes competitiveness. Games are competitive in nature when it involves people against each other. & 0.37 & Competitiveness also means education, research and innovation including in the area of environment. & -0.14\\
		\rule{0pt}{5ex}
		How is a dog likely to communicate with another dog? & bark & Bark is the sharp explosive cry of a dog, fox, or seal. The dog is likely to communicate with another dog with a bark. & 2.11 & Bark is covering. & -4.37\\
            \rule{0pt}{5ex}
		Where would you put a car near your house? & driveway & Driveway is a place near the house. A car can be put in the driveway. & 0.48 & Driveways are located in cars. & 0.43\\
		\bottomrule
	\end{tabular}
 }
	\caption{Exemplar of REV scores for crowdsourced rationales and label-related but input-irrelevant sentences containing the ground-truth label from GenericsKB for ECQA.}
	\label{tab:trivial_exp}
\end{table*}
In some cases, a rationale contains the given label and provides new information related to the label, but does not necessarily explain why the label is selected for the input. 
To evaluate such rationales, we randomly select 250 gold labels in ECQA and extract their related sentences from a large-scale knowledge base—GenericsKB \citep{bhakthavatsalam2020genericskb}. 
Those sentences contain the labels, while might provide little or irrelevant new information to explain the labels w.r.t. the inputs. 
We use them as trivial rationales for evaluation. 
The average \ourmetric scores for those trivial rationales and their crowdsourced counterparts are 0.26 and 1.14 respectively, indicating the effectiveness of \ourmetric in identifying the new and relevant information in rationales. 
Table \ref{tab:trivial_exp} shows the REV scores of some examples and the corresponding crowdsourced rationales. 
The results show that \ourmetric can distinguish the new information in different rationales and penalize meaningless rationales. 
Overall, \ourmetric gives higher scores to crowdsourced rationales than trivial sentences from GenericsKB. 

\subsection{Qualitative Analysis of CoS-E Rationales}
\label{sec:qualitative_cose}
Table \ref{tab:cose_exp} shows the exemplar of REV scores for crowdsourced and model-generated (\outputr) rationales for CoS-E. 
The main observation is model-generated rationales (\outputr) generally support labels, though provide limited new information, while many crowdsourced rationales in CoS-E are noisy or uninformative. Specifically, compared to the crowdsourced rationales in CoS-E, we observe that \outputr can produce better rationales that support the labels, which also corresponds to higher REV scores. However, the new information contained in those rationales is still limited (please see examples). A possible reason is the task model (\outputr) hardly learns to produce more informative rationales when trained using lower quality rationales from CoS-E, known quality issue as reported in prior work \citep{aggarwal-etal-2021-explanations, sun2022investigating}.

\subsection{Human Evaluation Details}
\label{sec:human_eval_supp}
We randomly select 230 examples from the ECQA test set and conduct human evaluation on the four types of rationale-label pairs (\gold, \outputr, \outputyr, \outputry) w.r.t. each example through the Amazon Mechanical Turk (AMT). 
We select workers located in Australia, Canada, the UK, or the US, with a past HIT approval rate of >98\% and >5000 HITs
approved. 
Each instance is assessed by 3 workers. 
We pay the workers $\$0.08$ for assessing each instance. 

\begin{figure*}[h]
  \centering
  \includegraphics[width=0.95\textwidth]{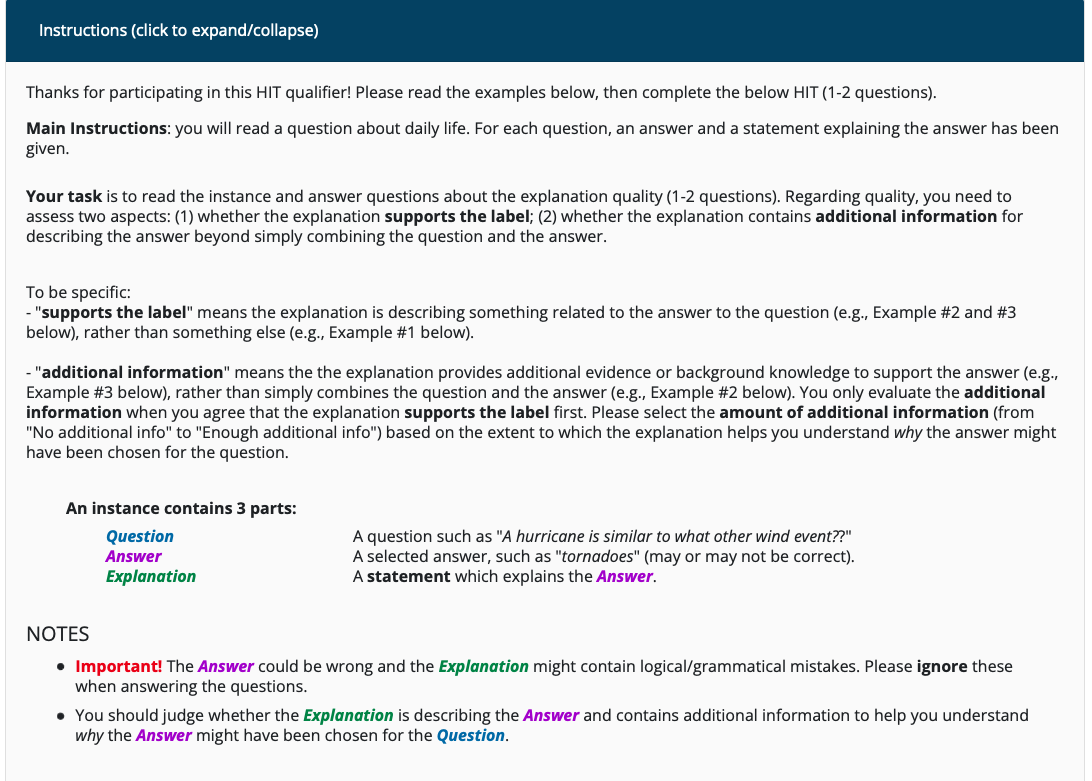}
  \caption{\label{fig:interface1}The instructions of human evaluation in the user interface on AMT.}
\end{figure*}

\begin{figure*}[h]
  \centering
  \includegraphics[width=0.95\textwidth]{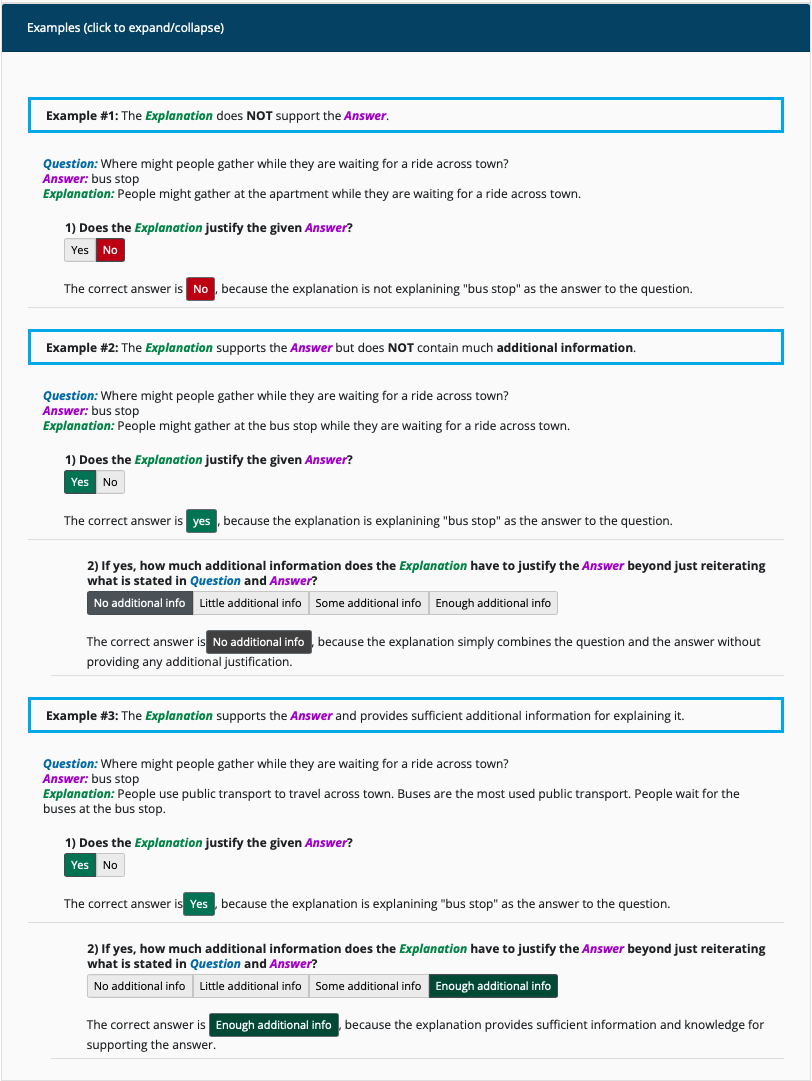}
  \caption{\label{fig:interface3}Exemplars provided to worker in the user interface on AMT.}
\end{figure*}

\begin{figure*}[h]
  \centering
  \includegraphics[width=0.95\textwidth]{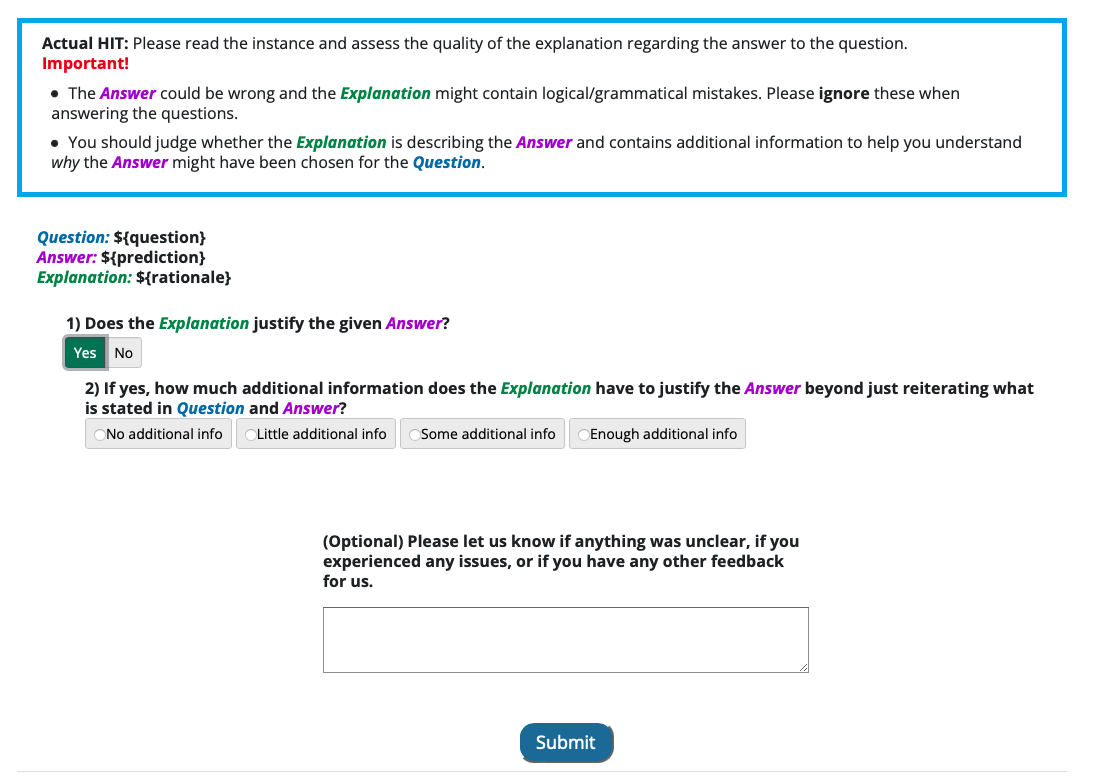}
  \caption{\label{fig:interface2}The actual hit of human evaluation in the user interface on AMT.}
\end{figure*}

Figure \ref{fig:interface1} shows the instructions we provide to workers. 
In Figure \ref{fig:interface3}, we show three examples, illustrating when the explanation (rationale) does not justify the answer (label), when the explanation supports the answer while not supplying additional information, and when the explanation supports the answer and provides additional information. 
Figure \ref{fig:interface2} shows the interface of the actual hit for human evaluation.

For each instance, we provide a question (input), an answer (label), and an explanation (rationale), and ask the workers to answer the following two questions:
\begin{enumerate}
    \item \textit{Does the Explanation justify the given Answer?} (yes or no) The question is to ask workers to judge whether the rationale supports the label or not.
    \item \textit{If yes, how much additional information does the Explanation have to justify the Answer beyond just reiterating what is stated in Question and Answer?} (No additional info, Little additional info, Some additional info, Enough additional info) We only ask this question if the workers choose ``yes'' for the first question. We design this question to ask workers to evaluate the extent to which the rationale provides additional information for justifying the label beyond repeating it w.r.t. the input. 
\end{enumerate}

\subsection{Qualitative Results of Sensitivity Test}
\label{sec:sensitivity_test_supp}
Table \ref{tab:sensitivity_test} shows some examples from the sensitivity test in \S\ref{sec:senitivity_test}. 
\begin{table*}[h]
\footnotesize
	\centering
	\begin{tabular}{p{13cm}}
		\toprule
		\textbf{\outputyr} \\
		\textit{Question}: What is likely to happen if you're doing housework all day? \\
		\textit{Choices}: boredom, tiredness, get tired, backache, get a clean house \\
		\textit{Gold label}: get tired \\
		\textit{Gold rationale}: Get tired means no longer wanting someone or wanting to do something because you are bored with it or annoyed by it. Get tired is likely to happen if you're doing housework all day. \\
		\rule{0pt}{3ex}
		$\mathbf{\sigma^{2}=0, \textbf{\ourmetric}=1.36}$ \\
		\textit{Output}: get tired [rationale] If you're doing housework all day, it's likely to happen if you are getting tired. \\
		\rule{0pt}{3ex}
		$\mathbf{\sigma^{2}=5, \textbf{\ourmetric}=1.36}$ \\
		\textit{Output}: get tired [rationale] If you're doing housework all day, it's likely to happen if you are getting tired. \\
		\rule{0pt}{3ex}
		$\mathbf{\sigma^{2}=10, \textbf{\ourmetric}=1.36}$ \\
		\textit{Output}: get tired [rationale] If you're doing housework all day, it is likely to happen if you are getting tired.\\
		\rule{0pt}{3ex}
		$\mathbf{\sigma^{2}=15, \textbf{\ourmetric}=1.29}$ \\
		\textit{Output}: backache [rationale] Backache is a condition of being tired. You are likely to get tired if you're doing housework all day. \\
		\rule{0pt}{3ex}
		$\mathbf{\sigma^{2}=20, \textbf{\ourmetric}=0.72}$ \\
		\textit{Output}: boredom [rationale] Boredom is a state of being or being unable to do something. People tend to get bored easily when they are bored. \\
		\rule{0pt}{3ex}
		$\mathbf{\sigma^{2}=25, \textbf{\ourmetric}=1.33}$ \\
		\textit{Output}: backache [rationale] Backache is the state of being unable to move backwards or forwards in a straight line. One cannot move forward if they are not able to do so. \\
		\rule{0pt}{3ex}
		$\mathbf{\sigma^{2}=30, \textbf{\ourmetric}=0}$ \\
		\textit{Output}: banishing oneself from one's own body by removing it from the body of another person. This is a way of preventing one from being harmed by other people. It is possible to get rid of people from their own bodies if they are living in the same house. That is why they have to live together. \\
		\midrule
		\textbf{\outputry} \\
		\textit{Question}: What happens when not paying attention when bathing? \\
		\textit{Choices}: dry skin, using soap, reading newspaper, shampoo, slip and fall \\
		\textit{Gold label}: slip and fall \\
		\textit{Gold rationale}: Bathing is generally done in a bathroom. Bathrooms could be slippery due to water. If not careful slipperiness can cause one to slip and fall. \\
		\rule{0pt}{3ex}
		$\mathbf{\sigma^{2}=0, \textbf{\ourmetric}=1.28}$ \\
		\textit{Output}: Bathing can cause slip and fall when one is not paying attention. Slip and Fall is a common occurrence when people do not pay attention when bathing. [answer] slip and fall \\
		\rule{0pt}{3ex}
		$\mathbf{\sigma^{2}=5, \textbf{\ourmetric}=1.28}$ \\
		\textit{Output}: Bathing can cause slip and fall when one is not paying attention. Slip and Fall is a common occurrence when people do not pay attention when bathing. [answer] slip and fall \\
		\rule{0pt}{3ex}
		$\mathbf{\sigma^{2}=10, \textbf{\ourmetric}=1.24}$ \\
		\textit{Output}: Bathing can cause slip and fall when not paying attention. Slipping and falling is a result of not being attentive when bathing. [answer] slip and fall  \\
		\rule{0pt}{3ex}
		$\mathbf{\sigma^{2}=15, \textbf{\ourmetric}=1.20}$ \\
		\textit{Output}: Bathing is a process of washing the body. Not paying attention means not paying enough attention to the bathing process. [answer] slip and fall \\
		\rule{0pt}{3ex}
		$\mathbf{\sigma^{2}=20, \textbf{\ourmetric}=-0.15}$ \\
		\textit{Output}: Bathing is a process of washing and rinsing. Paying attention means paying attention to something. [answer] reading newspaper \\
		\rule{0pt}{3ex}
		$\mathbf{\sigma^{2}=25, \textbf{\ourmetric}=-0.17}$ \\
		\textit{Output}: Paying attention to what you are doing is a sign of being aware of what is going on around you. Knowing what to do is to pay attention. [answer] dry skin \\
		\rule{0pt}{3ex}
		$\mathbf{\sigma^{2}=30, \textbf{\ourmetric}=-0.85}$ \\
		\textit{Output}: Paying attention to one's own needs is a basic human need. One needs to be aware of the needs of others in order to meet them...engavengendersengendoenv -end-engagingendentenant [answer] slip and fall \\
		\bottomrule
	\end{tabular}
	\caption{Examples of the two task models' outputs under different levels of noise in sensitivity test.}
	\label{tab:sensitivity_test}
\end{table*}

\begin{table*}[tbh]
\small
	\centering
	\begin{tabular}{P{2cm}P{4cm}P{2cm}P{3.5cm}P{0.8cm}}
		\toprule
		Type & Input & Label & Rationale & REV \\
		\midrule
            \multirow{3}{*}{Crowdsourced}  & The goal was to hit the target, but a projectile ball can't hit anything if it isn't in what? & motion & if you stand still you get hit & -0.14 \\
		\rule{0pt}{5ex}
		& When you get together with friends to watch film, you might do plenty of this? & have fun & when the working day is done & -0.27 \\
		\rule{0pt}{5ex}
		& They dealt with combustible mixtures in their experiments, this is why they kept a fire extinguisher where? & chemistry lab & mixtures mixing fruitsa & -0.17 \\
		\midrule
		\multirow{3}{*}{$XY^* \rightarrow R$} & The goal was to hit the target, but a projectile ball can't hit anything if it isn't in what? & motion & a projectile ball can't hit anything if it's not in motion & 0.09 \\
		\rule{0pt}{5ex}
		& When you get together with friends to watch film, you might do plenty of this? & have fun & when you get together with friends to watch film, you might do plenty of fun & 1.47 \\
		\rule{0pt}{5ex}
		& They dealt with combustible mixtures in their experiments, this is why they kept a fire extinguisher where? & chemistry lab & chemistry labs deal with combustible mixtures in their experiments. & 0.74 \\
		\bottomrule
	\end{tabular}
	\caption{Exemplar of REV scores for crowdsourced and model-generated (\outputr) rationales for CoS-E.}
	\label{tab:cose_exp}
\end{table*}

\begin{table*}[tbh]
\small
	\centering
	\begin{tabular}{P{4cm}P{2cm}P{4cm}P{0.8cm}}
		\toprule
	    Input & Label & Rationale & \ourmetric \\
		\midrule
		What do people call it when they are going for run? & falling down & People call it run when they are going for run. & -1.06 \\
		\rule{0pt}{5ex}
		What enables most people to transport themselves? & own cars & People who believe in god are able to transport themselves through helicopter. & -0.19 \\
		\rule{0pt}{5ex}
		Where would you expect to find popcorn in a public place? & movie theater & Popcorn can be found in a public place. & -1.27 \\
		\rule{0pt}{5ex}
		What are you usually at when you sit on a bench on a curb? & city & Ohio is a state in the United States. You are usually at street corner when you sit on bench on curb. & -0.27 \\
		\bottomrule
	\end{tabular}
	\caption{Exemplar of negative REV scores for rationale-label pairs from \outputry on the ECQA dataset.}
	\label{tab:neg_rev}
\end{table*}

\end{document}